\documentclass[preprint,authoryear,a4paper]{elsarticle}
\usepackage{verbatim}

\usepackage{geometry}

\usepackage{color,soul}

\usepackage[hidelinks]{hyperref}
\hypersetup{
  colorlinks   = true, 
  urlcolor     = blue,
  linkcolor    = blue, 
  citecolor    = blue 
}

\usepackage{graphicx}
\graphicspath{{./}}
\DeclareGraphicsExtensions{.pdf,.png,.jpg}
\usepackage{placeins}
\usepackage{caption}
\usepackage{subcaption}

\usepackage{booktabs}
\usepackage[table,xcdraw]{xcolor}
\usepackage{graphicx}
\usepackage{adjustbox}
\usepackage{multirow}
\usepackage{diagbox}

\usepackage{amssymb}
\usepackage{gensymb}
\usepackage[mathscr]{euscript}
\usepackage{bm}
\usepackage{mathtools}
\usepackage{nccmath}
\usepackage{amsmath}
\usepackage{braket}

\usepackage{algorithm}
\usepackage{algpseudocode}

\usepackage{lineno, hyperref}

\usepackage{lipsum}

\usepackage{ulem}

\journal{ISPRS Journal of Photogrammetry and Remote Sensing}

\begin{document}

\begin{frontmatter}

    \title{Semantic Image Translation for Repairing the Texture Defects of Building Models}

    \author{Qisen Shang}
    \author{Han Hu\corref{cor1}}
    \author{Haojia Yu}
    \author{Bo Xu}
    \author{Libin Wang}
    \author{Qing Zhu}
    \cortext[cor1]{Corresponding Author: han.hu@swjtu.edu.cn}

    \address{Faculty of Geosciences and Environmental Engineering, Southwest Jiaotong University, Chengdu, China}

    \begin{abstract}     
        The accurate representation of 3D building models in urban environments is significantly hindered by challenges such as texture occlusion, blurring, and missing details, which are difficult to mitigate through standard photogrammetric texture mapping pipelines. Current image completion methods often struggle to produce structured results and effectively handle the intricate nature of highly-structured fa\c{c}ade textures with diverse architectural styles. Furthermore, existing image synthesis methods encounter difficulties in preserving high-frequency details and artificial regular structures, which are essential for achieving realistic fa\c{c}ade texture synthesis.
        To address these challenges, we introduce a novel approach for synthesizing fa\c{c}ade texture images that authentically reflect the architectural style from a structured label map, guided by a ground-truth fa\c{c}ade image. In order to preserve fine details and regular structures, we propose a regularity-aware multi-domain method that capitalizes on frequency information and corner maps. We also incorporate SEAN blocks into our generator to enable versatile style transfer. To generate plausible structured images without undesirable regions, we employ image completion techniques to remove occlusions according to semantics prior to image inference.
        Our proposed method is also capable of synthesizing texture images with specific styles for fa\c{c}ades that lack pre-existing textures, using manually annotated labels. Experimental results on publicly available fa\c{c}ade image and 3D model datasets demonstrate that our method yields superior results and effectively addresses issues associated with flawed textures. The code and datasets will be made publicly available for further research and development.
    \end{abstract}

\end{frontmatter}

\section{Introduction}
\label{s:intro}
\begin{figure}[H]
	\centering
	\subcaptionbox{Occlussions of SWJTU model.}[0.53\linewidth]{
		\includegraphics[width=\linewidth]{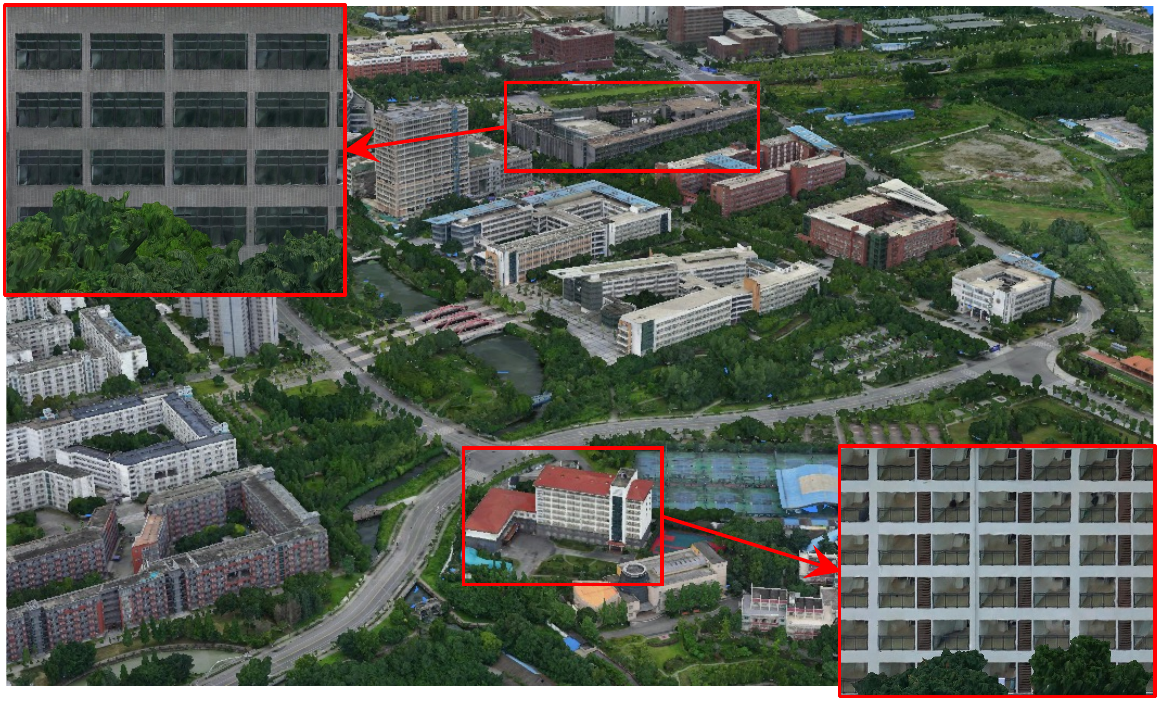}
	}
	\subcaptionbox{Texture missing of London model.}[0.46\linewidth]{
		\includegraphics[width=\linewidth]{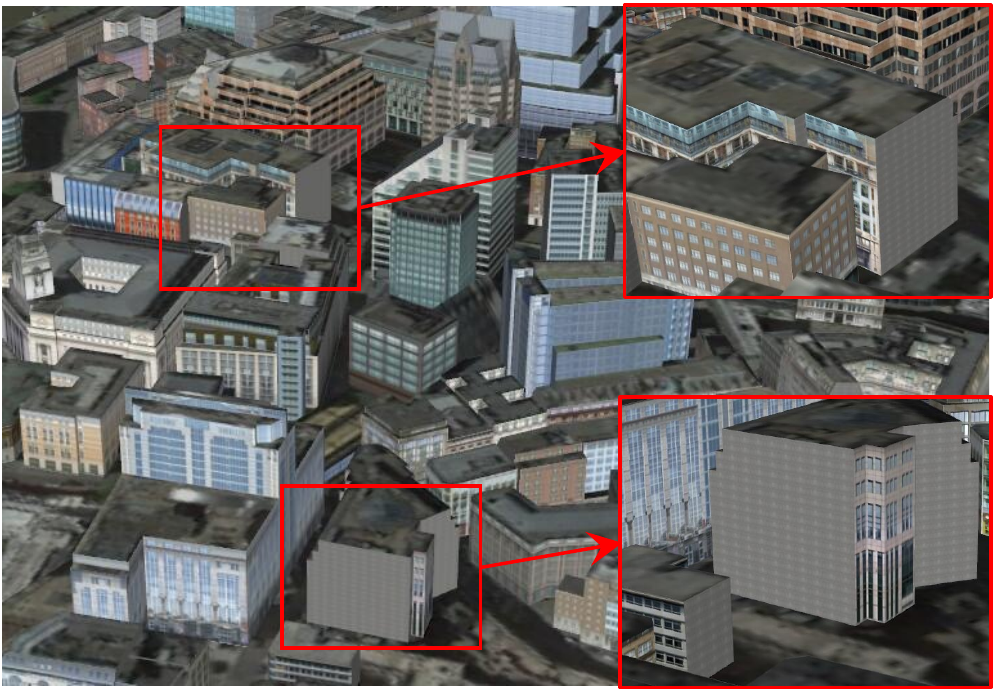}
	}
	\caption{Defective textures of 3D building models. The red rectangles denote enlarged regions.}
	\label{fig:defective_texture}
\end{figure}

Three-dimensional (3D) building models are fundamental to the development of digital cities, serving as a crucial component in high-precision mapping, autonomous driving, and urban planning \citep{lin2013virtual, tao2019make}. The realism of 3D models is conveyed through their geometry and texture \citep{chen2020combining, buyukdemircioglu2020reconstruction}. At present, high-precision building models are predominantly created manually, while textures are primarily sourced from various imagery.
However, in densely populated urban areas, aerial imagery is often obstructed by building occlusions, and ground-level photography is impeded by adjacent objects, such as vegetation and billboards. As a result, acquiring unobstructed imagery from any angle and distance proves challenging, rendering traditional texture mapping pipelines inadequate for processing or de-occluding building fa\c{c}ade images \citep{zhou2020selection, zhu2021depth, zhang2021continuous, li2023optimized}. As depicted in Figure \ref{fig:defective_texture}, defective textures are prevalent in various realistic building models.

Current models frequently utilize occluded textures directly or substitute them with manually crafted repetitive textures. In certain cases, textures are selected from pre-existing material libraries. These processing techniques substantially constrain the visualization quality of models \citep{li2023optimized}. To improve the visual fidelity of realistic 3D building models, it is crucial to tackle the issue of defective textures. As a result, this paper concentrates on repairing defective textures in realistic 3D building models to facilitate high-precision urban 3D reconstruction and advanced applications. Although there have been advancements in texture processing and occlusion removal, a number of outstanding challenges still need to be addressed.

\textit{1) Inaccessible fa\c{c}ade textures in built-up areas.}
Multi-view aerial camera systems, mobile measurement systems (MMS), and handheld cameras are widely employed and effectively integrated in urban realistic 3D modeling for tasks such as feature matching and texture mapping \citep{remondino2015oblique, zhu2020leveraging, zhu2021structure}. However, there remain many areas that are inaccessible to these sensors, particularly buildings in built-up areas. Some researchers have attempted to mitigate occlusions by selecting optimal pixels from various aerial images for texture remapping, but this approach is limited to areas that are not fully occluded \citep{zhou2020selection, yang2021moving}. A mesh completion method has been proposed in previous work, which processes 3D textures in 2D space and employs image completion methods to repair textures in occluded road areas. Nevertheless, this method cannot handle unseen pixels or generate textures from scratch, which is a frequent issue in built-up areas \citep{zhu2021structure}. Moreover, we observe that recovering the semantics of occluded buildings is considerably more manageable than completing textures. As such, synthesizing textures from recovered or specified semantic labels presents a practical solution.

\textit{2) Highly structured and diverse styles of building fa\c{c}ades.}
Windows, doors, balconies, and other components on fa\c{c}ades contribute to the textures of buildings. An ideal building fa\c{c}ade texture should exhibit structured man-made components and a realistic style with details. However, neither data-driven nor patch-based methods possess sufficient generalization for different buildings with various styles, and they cannot generate highly structured de-occluded or synthesized textures \citep{criminisi2004region, zhu2021structure}. Moreover, the results of existing image synthesis methods lack texture details, such as bricks and window frames, which are crucial for building fa\c{c}ades \citep{cai2021frequency}. The regularization and generalization of current methods cannot fulfill the requirements of building fa\c{c}ade texture repair. As such, one viable solution is to first recover or provide the semantics of fa\c{c}ades to control regularization, and then translate the semantic label map into textures by combining patch-based and data-based methods.

To address the issues outlined above, we propose a method for repairing building fa\c{c}ade textures to enhance the realism of 3D building models. Our method involves synthesizing realistic fa\c{c}ade textures by using unoccluded semantic labels and ground-truth fa\c{c}ade images, a process also known as image translation \citep{isola2017image}. Our approach is capable of handling both occluded and missing fa\c{c}ade textures. The difference in processing occlusion and missing fa\c{c}ade textures lies primarily in the acquisition of semantic labels. For occlusion, we use an image completion algorithm to recover the semantics of occluded regions. For missing textures, we manually annotate the semantics of the corresponding fa\c{c}ade. We then utilize the semantic labels as content and the ground-truth fa\c{c}ade images as style to train a generative adversarial network (GAN) to synthesize pseudo fa\c{c}ade textures \citep{goodfellow2020generative}. If the wall style is too structured to be synthesized by the GAN, our method falls back on image quilting to generate more regular results \citep{efros2001image}.

In summary, this paper offers the following two contributions to repairing defective textures of building fa\c{c}ades: 1) a practical solution for texture occlusion or missing building fa\c{c}ades through image translation from unoccluded semantic labels, and 2) a novel arbitrary label-to-image translation method with rich details and regular structures. The rest of this paper is organized as follows: Section 2 provides a brief review of related work. Section \ref{s:method_overview} introduces the workflow of the proposed fa\c{c}ade image synthesis approach. Sections \ref{s:semantic_completion}, \ref{subs:image_translation}, and \ref{subs:image_quilting} elaborate on the details of the proposed method. Experimental evaluations are presented in Section \ref{s:results}. Finally, Section \ref{s:conclusion} concludes the paper.

\section{Related work}
\label{s:related_work}

In the following, we only discuss the most relevant literature, including 1) texture mapping and de-occlusion, 2) semantic recovery and 3) image translation.

\paragraph{1) Texture mapping and de-occlusion}
Significant progress has been made in bundle adjustment \citep{verykokou2018oblique} and dense image matching \citep{hirschmuller2007stereo, hu2016stable}, enabling realistic urban modeling. To further advance applications such as autopilot, researchers have obtained monolithic building models using parametric \citep{kelly2017bigsur, kelly2018frankengan} or interactive modeling \citep{vanegas2012procedural, kelly2011interactive} approaches. These models achieve realism through texture mapping of images acquired from different platforms. The concept of texture mapping was first proposed by \cite{catmull1974subdivision}, and subsequent studies by \cite{sinha2008interactive} and \cite{tan2008large} realized texture mapping for 3D models through interactive approaches.
One approach to automated texture mapping is to use a Markov Random Field (MRF) energy function to select optimal images for each facet, as demonstrated by \cite{lempitsky2007seamless} and \cite{waechter2014let}. \cite{gal2010seamless} further improved upon this method by introducing clarity measurement and translation vectors to achieve a smoother textured mesh. However, these single-view-based methods have limitations in terms of texture alignment with geometric structures.
In contrast, multi-view-based methods fuse multiple images to obtain more consistent textures for each facet \citep{callieri2008masked, grammatikopoulos2007automatic}. However, these methods have strict requirements on reconstruction accuracy and image resolution, and their low efficiency limits their widespread application \citep{waechter2014let}.

Removing occlusions for 3D models with discontinuous textures is a challenging problem. While \cite{grammatikopoulos2007automatic} addressed this issue by automatically filtering out texture outliers using statistical tests, and \cite{yu2019free, yang2021moving} used deep-learning-based target detection methods to detect occlusions and remap textures to eliminate some of them, these methods cannot solve the inherent problem embedded in the texture mapping pipeline of multi-view images, which lacks imagination for invisible areas \citep{zhu2021structure}.
Our previous work \citep{zhu2021structure} successfully solved this problem for road areas using offscreen rendering and image completion, but this method has limitations when applied to building fa\c{c}ades. Specifically, it cannot generate highly structured textures like those found in building fa\c{c}ades and lacks generalization ability for different architectural styles, e.g., Bauhaus and Baroque \citep{zhu2021structure}. To overcome these limitations, we propose a practical approach to de-occlusion by synthesizing texture from semantic labels and ground-truth images using GAN \citep{goodfellow2020generative}. Here, semantic labels control the structure of synthetic content, while ground-truth images specify the style \citep{isola2017image}.

\paragraph{2) Semantic recovery}
Building fa\c{c}ades always exhibit a highly structured character, which was exploited by \cite{stiny1975pictorial, ripperda2009application} to parse building fa\c{c}ades for reconstruction. \cite{koutsourakis2009single} proposed a fa\c{c}ade parsing method guided by MRF, and subsequent work by \cite{teboul2011shape} and \cite{cao2017facade} improved upon this method to achieve better results with higher efficiency. While this approach is less susceptible to occlusion, it is also complex and difficult to apply to different architectural styles. To address this limitation, researchers have turned to supervised learning methods for fa\c{c}ade parsing using labeled data \citep{martinovic2013bayesian, gadde2017efficient, dehbi2017statistical}. However, these methods struggle with the complexity of fa\c{c}ades with varying styles. To alleviate this issue, researchers have explored the use of repeating patterns in buildings \citep{muller2007image, friedman2012online, zhang2013layered, fan2014structure, cohen2017symmetry}.

In recent years, deep learning-based object detection has revolutionized the field of computer vision, with R-CNN \citep{girshick2014rich} being a pioneering method. Mask R-CNN \citep{he2017mask} extended Faster R-CNN \citep{ren2015faster} by adding a mask prediction branch, enabling semantic segmentation. While these methods have shown promising results in building fa\c{c}ade annotation, they are still limited by occlusions. To address this issue, \cite{lin2019fusion} and \cite{hu2020fast} utilized multi-source data, such as infrared or panorama data, to alleviate occlusion using semantic segmentation methods. However, the cost of acquiring multi-source data is often high, and there may still be areas that are inaccessible to sensors.

We have observed that most de-occlusion requirements are for reconstructed textured models. Semantic labels can be easily and accurately captured by existing methods or through manual annotation, and the semantics of occlusions are easy to determine. Therefore, we propose generating masks based on label maps, and then recovering the masked regions using image completion to achieve the goal of de-occlusion.

Image inpainting and completion techniques are commonly used to fill in missing or undesirable parts of an image with plausible pixels. Traditional methods for this task include partial differential equations (PDEs) and sampling-based approaches \citep{bertalmio2000image, criminisi2004region}. PDE-based approaches lack attention to global information and cannot fill large holes \citep{zhu2021structure}. On the other hand, sampling-based approaches fill in the void regions by using global similar patches, which are translated and rotated, and can repair large regions. The patch matching algorithm proposed by \cite{barnes2009patchmatch} significantly accelerated the search for similar patches and made sampling-based methods state-of-the-art. Subsequently, \cite{he2012statistics}, \cite{huang2014image}, and \cite{zhu2021structure} improved the patch-match based methods by incorporating offsets statistics, affine deformation, and linear patterns, respectively. While GAN-based approaches have shown impressive results on benchmark datasets, they rely on massive labeled data such as the ffhq-dataset \citep{karras2019style}. However, unlike other easily accessible datasets, building fa\c{c}ade images require complex processing such as geometric deformation correction, making it difficult to obtain massive training data \citep{zhu2020large}. Moreover, the complexity of semantic labels for building fa\c{c}ades is much lower than that of photorealistic images. Therefore, in this paper, we choose the patch-match based algorithm for the semantic recovery of occluded regions.

\paragraph{3) Image translation}
GANs employ an adversarial strategy to train two networks: a generator that simulates the probability distribution of training data from random signals and a discriminator that discerns whether the generated samples are real or fake \citep{goodfellow2020generative}. Unlike CNNs, GANs use a zero-sum game between the generator and discriminator to reach Nash equilibrium, thereby enhancing the generator's ability \citep{goodfellow2020generative}. Vanilla GAN generates sharper samples from random signals than Variational Auto Encoder (VAE) \citep{pu2016variational}, paving the way for a new image synthesis approach \citep{goodfellow2020generative}. DCGAN introduces CNN structure for stable training \citep{radford2015unsupervised}. WGAN replaces the Kullback-Leibler (KL) and Jensen-Shannon (JS) divergence with Wasserstein distance for measurement, solving the vanishing gradient problem \citep{arjovsky2017wasserstein}. LSGAN sets the objective function to the squared difference form, resulting in a more stable training process and better results \citep{mao2017least}. PGAN adopts a progressive training strategy to generate higher-resolution images \citep{karras2017progressive}.

Controlling the inference process can be challenging since GANs generate results from random input signals. Conditional GANs, designed to address this issue, modify the random inputs into conditional maps \citep{mirza2014conditional}. In this paper, semantic labels are used as the conditional maps. Pix2Pix introduces PatchGAN, which evaluates generated results with a patch-based discriminator \citep{isola2017image}. Additionally, \cite{isola2017image} designs a U-Net \citep{ronneberger2015u} generator, improving performance on several benchmark datasets. Pix2PixHD \citep{wang2018high} builds on Pix2Pix by designing a multi-scale network to generate higher resolution images and adding perceptual \citep{johnson2016perceptual} and feature matching \citep{wang2018high} loss to control the style of generated images by learning in the latent space. The network is also trained with boundary maps to obtain clearer results.

For generating realistic and plausible fa\c{c}ade images, consistency with real images at a high level and accurate capture of low-level details are crucial. Style transfer is useful in this context as it adapts the style of a content image to match another image's style, acting as a form of domain adaptation for a single image \citep{jing2019neural}. Early style transfer methods, such as the one proposed by \cite{gatys2016image}, used deep features extracted through a DCNN and represented content and style using the Mean Squared Error (MSE) of the feature map and its Gram Matrix. However, this method only supported single style transfer.
To address this limitation, StyleBank \citep{chen2017stylebank} was developed to train multiple styles simultaneously, but it still couldn't transfer inputs to arbitrary styles outside the training dataset. Instance Normalization (IN) \citep{ulyanov2016instance} and Adaptive Instance Normalization (AdaIN) \citep{huang2017arbitrary} were introduced to enable arbitrary style transfer. StyleGAN \citep{karras2019style} built on this idea by injecting style information as AdaIN into the network, achieving impressive results on face datasets. Another approach by \cite{li2018closed} used a whitening and coloring transform for arbitrary style transfer, operating in the feature space extracted by a pre-training network (encoder) and requiring only a few reconstruction networks (decoder) for good transfer results.

In recent years, the combination of image translation and style transfer has seen remarkable results. SPADE (SPatially-Adaptive DE-normalization) \citep{park2019semantic} treated style transfer as a process of de-normalization and designed a SPADE ResBlk module to replace the ResBlk in Pix2PixHD \citep{wang2018high}, injecting style information extracted by a trainable encoder. SEAN \citep{zhu2020sean} improved SPADE by performing style transfer separately for different classes, achieving state-of-the-art performance. In this paper, we consider frequency and regular structure information to enhance the texture details and regularity of SEAN, meeting the fa\c{c}ade texture mapping needs of photorealistic building models. However, GANs may not be effective on all datasets due to their data-based implicit probability density estimation nature. To address this issue, the proposed algorithm falls back to image quilting in regions annotated as walls when deep learning methods are ineffective \citep{efros2001image}.

\section{Structured realistic image synthesis method for building fa\c{c}ades}
\label{approach}
\subsection{Overview and problem setup}
\label{s:method_overview}
\subsubsection{Overview of the approach}

\begin{figure}[H]
	\centering
	\includegraphics[width=\linewidth]{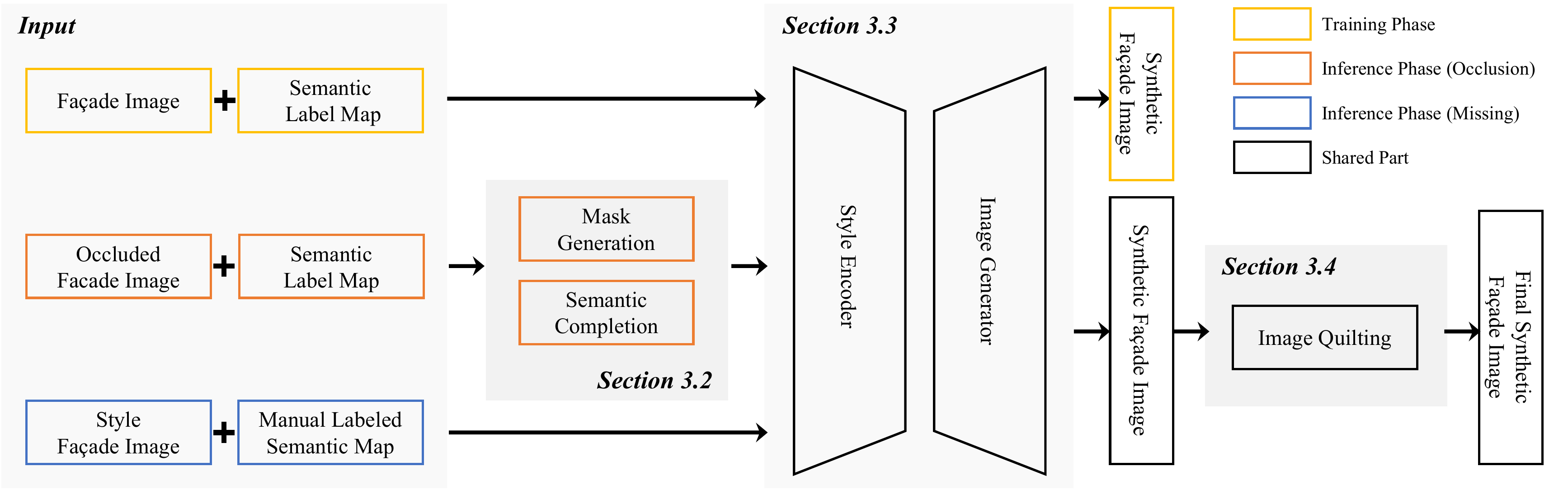}
	\caption{Workflow of proposed approach. Rectangles with different colors denote different phases, i.e. training or inference, and black rectangles denote the shared parts of these phases.}
	\label{fig:workflow}
\end{figure}

To address the challenges of building fa\c{c}ade texture occlusion and missing problems in built-up areas, we develop a label-to-image deep neural network that utilizes the ground-truth image as an additional input for style. This approach enables the generation of realistic fa\c{c}ade texture images from complete semantic labels. In addition to employing universal losses, such as GAN loss, L1 loss, and perceptual loss, we propose regularity loss and detail loss from multiple domains to enhance the regularity and detail of the results. Furthermore, we apply image completion to recover occluded semantics and use image quilting when the generation capacity of the trained network is insufficient. The overall workflow, consisting of training and inference phases, is illustrated in Figure \ref{fig:workflow}.
\paragraph{Training}
During each step of the training phase, we use a ground-truth image and its corresponding semantic label as inputs, providing style and content information, respectively. A trainable style encoder processes these inputs, extracting style vectors that represent different label styles. We then feed the semantic label and style vectors into a trainable image generator to synthesize a stylized image. To enhance the regularity and detail of the generated images, we transform both input and output images into frequency maps, spectrum maps, and corner maps. We calculate regularity loss and detail loss separately based on these maps during the back-propagation optimization. As in vanilla GANs, we also train a discriminator to compete with the generator \citep{goodfellow2020generative}.
\paragraph{Inference}
During the inference phase, our approach can address both texture occlusion and missing problems, with the primary differences arising from the acquisition of semantic labels fed into the image generator. For the occlusion problem, we identify occluded regions based on semantics and apply an image completion algorithm to recover semantics. For the missing problem, we manually provide a semantic label map to specify the content of the synthetic image. In this paper, the image used to specify style is set to an occluded fa\c{c}ade image for the occlusion problem and a desirable style image for the missing problem. Following a similar process to the first half of the training phase, we feed de-occluded or manually annotated semantic maps and style vectors into the trained image generator to synthesize realistic fa\c{c}ade textures with the actual style. Finally, if the generated result is not satisfactory, our approach employs image quilting to composite wall textures, improving the overall quality of the final result.

\subsubsection{Problem setup}
\label{subsubs:problem_setup}
As illustrated in Figure \ref{fig:workflow}, our approach consists of three trainable components: the style encoder $\boldsymbol{E}$, the generator $\boldsymbol{G}$, and the discriminator $\boldsymbol{D}$. The overall objective of our study is more formally presented in Equation \ref{equ:overall}. 
 \begin{equation}
 	\label{equ:overall}
 	\min\mathcal{V}( \mathbf{R'},  \mathbf{R})
 \end{equation}
where $\mathcal{V}$ denotes the measurement between two samples. $\mathbf{R'}\in\mathbb{R}^{H \times W \times 3}$ and $\mathbf{R}\in\mathbb{R}^{H \times W \times 3}$ denote the generated image and ground-truth image respectively. $\mathbf{R'}$ can be expressed as $\boldsymbol{G}\left( \mathbf{S},  \mathbf{M}^{\prime}\right)$, where $ \mathbf{S}\in\mathbb{R}^{H \times W \times 3\times N}$ is the style vectors calculated by equation $ \mathbf{S}=\boldsymbol{E}( \mathbf{R}, \mathbf{M})$. In $\boldsymbol{E}( \mathbf{R}, \mathbf{M})$, $\mathbf{R}$ and $\mathbf{M}\in\mathbb{B}^{H \times W \times 3}$ are the most primitive inputs of network, specifically, the ground-truth sample and its corresponding semantic label. $\mathbf{M'}\in\mathbb{B}^{H \times W \times 3}$ is the inputted semantic label map of generator $\boldsymbol{G}$ ($\mathbf{M'} = \mathbf{M}$ in training phase).

\subsection{Direction guided semantics completion}
\label{s:semantic_completion}
\begin{figure}[h]
	\centering
	\includegraphics[width=\linewidth]{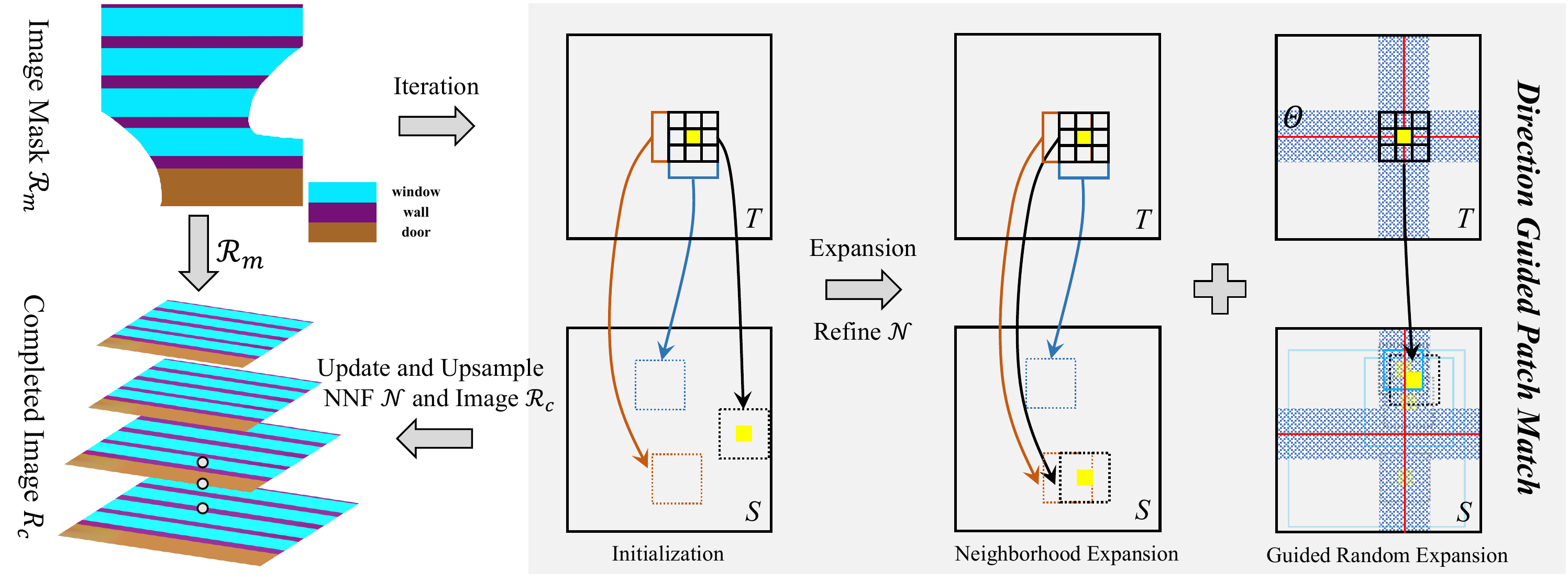}
	\caption{Workflow of direction guided semantics completion.}
	\label{fig:semantic_completion}
\end{figure}

Due to the difficulty in obtaining structured results through direct image completion, the semantic label map of a building fa\c{c}ade, particularly manually labeled semantic maps, provides a highly structured alternative that can be easily used to synthesize a desirable fa\c{c}ade image. However, to achieve de-occlusion through image translation from the label map, the first challenge is semantics recovery. We identify the occluded regions in the label map based on predefined semantics, and then intuitively employ an image restoration method for semantics completion. As label maps are simpler than natural images and data-driven methods are excessive, we opt for a patch-based algorithm to accomplish our image restoration goal. The pipeline of the proposed semantics completion is illustrated in Figure \ref{fig:semantic_completion}.

Our objective for semantics completion is to recover the pixels of the void region in $\mathcal{R}_m$, which is generated from occluded or missing semantics. The algorithm establishes a pyramid to progressively complete the image by searching for the best nearest neighbor field (NNF) $\mathcal{N}$. The algorithm adopts a scanline-based expansion strategy in the patch match of each level \citep{barnes2009patchmatch}. For each pixel $\boldsymbol{p}$ in $\mathcal{R}_m$, we optimize the NNF $\mathcal{N}(\boldsymbol{p})$ using an improved strategy guided by the direction $\Theta$ \citep{zhu2021structure}.

\subsubsection{Similarity measure of patches}
For efficiency without sacrificing accuracy, the vanilla patch-match algorithm compares the similarity between the current pixel $\boldsymbol{p}$ and its four neighbors at offset $\boldsymbol{v}=\mathcal{N}(\boldsymbol{p})$. More specifically,
\begin{equation}
	\begin{split}
		T(\boldsymbol{p})&=\{\mathcal{R}_c(\boldsymbol{p}+\boldsymbol{s})|\boldsymbol{s}\in[-\tfrac{W}{2},\tfrac{W}{2}]\times[-\tfrac{W}{2},\tfrac{W}{2}] \} \\
		S(\boldsymbol{p}, \boldsymbol{v})&=T(\boldsymbol{p}+\boldsymbol{v})
	\end{split}
\end{equation}
where $T$ and $S$ are pixel patches centered on $\boldsymbol{p}$ and $\boldsymbol{p}+\boldsymbol{v}$ in the target and source domains, respectively. The target domain and the source domain correspond to the void and known regions in $\mathcal{R}_m$, as well as the occluded and unoccluded regions in $\mathcal{R}$. During the random expansion introduced in \ref{subsubs:expansion}, $\boldsymbol{v}$ can be either $\mathcal{N}(\boldsymbol{p})$ or its four neighbors $\mathcal{N}_4(\boldsymbol{p})$. The input label map size in this paper is 256 $\times$ 256, and in order to balance effect and efficiency, we set the patch-size $W$ to 7. Based on our previous research, the measurement $E$ of pixel similarity in this paper is shown in equation \ref{equ:E}.
\begin{equation}
	\label{equ:E}
	E=E_a+\lambda_1 E_p + \lambda_2 E_d
\end{equation}
The terms $E_a$, $E_p$, and $E_d$ represent appearance, proximity, and direction costs, respectively, and will be detailed in the following paragraphs. Additionally, $\lambda_1$ and $\lambda_2$ are empirically set to $5\times10^{-4}$ and $0.5$, respectively.

\paragraph{1) Appearance cost}
We measure the appearance similarity between patches using the following equation \ref{equ:Ea} which computes a Gaussian weighted sum of the absolute value difference. Here, $w_i$ is an isotropic weight generated from a Gaussian kernel \citep{huang2014image}.
\begin{equation}
	\label{equ:Ea}
	E_a(\boldsymbol{p},\boldsymbol{v})=\sum_{i} w_i| T_i(\boldsymbol{p}) - T_i(\boldsymbol{p}+\boldsymbol{v}))|
\end{equation}

\paragraph{2) Proximity cost}
Researchers have shown that better pixels tend to appear in closer patches. Proximity cost is used to penalize the selection of nearby pixels and is shown in equation \ref{equ:Ep}.
\begin{equation}
	\label{equ:Ep}
	E_p(\boldsymbol{p,v})=\frac{||\boldsymbol{v}||^2}{\sigma_d(\boldsymbol{p})^2+\sigma_c^2}
\end{equation}
where $\sigma_d(\cdot)$ calculate the minimum distance from the current pixel to the void region boundary, and $\sigma_c=\max(w,h)/8$ \citep{zhu2021structure}.

\paragraph{3) Direction cost}
The building fa\c{c}ade textures are corrected orthophotos with horizontally or vertically distributed components. Thus, we can utilize this pattern to evaluate the selected pixels for better results. This can be formulated as the following equation:
\begin{equation}
	E_r(\boldsymbol{v}) = \min_{\theta \in \Theta}{\cos(\theta_{\boldsymbol{v}} - \theta)}
\end{equation}
where $\theta_{\boldsymbol{v}}$ denotes the direction of current offset $\boldsymbol{v}$, $\theta$ is the element of $\Theta =\{\boldsymbol{\pi/2}, \boldsymbol{\pi}\}$.

\subsubsection{Direction guided expansion}
\label{subsubs:expansion}
The optimization of the NNF $\mathcal{N}(\boldsymbol{p})$ is an iterative process of random expansion. In a single iteration, for every pixel $\boldsymbol{p}\in \Omega$, we compare $E(\boldsymbol{p}, \mathcal{N}(\boldsymbol{p}))$ to $E(\boldsymbol{p}, \mathcal{N}(\boldsymbol{q}))$ twice. The first time, $\boldsymbol{q}$ comes from $N_4(\boldsymbol{p})$, which are the four neighbors of the current pixel $\boldsymbol{p}$. Subsequently, $\boldsymbol{q}$ comes from a set of random pixels $R(\boldsymbol{p})$. The elements of $R(\boldsymbol{p})$ are selected from the pixels distributed in the known regions within a radius $r$ centered on $\boldsymbol{p}$. The value of $r$ is initialized with $\max(w,h)$ and is halved per iteration until it reaches $1$ or other predetermined stopping conditions. Additionally, we constrain the areas of random expansion by the rectangular buffer generated along the direction $\Theta$, as we did in our previous work \citep{zhu2021structure}.

\subsection{Regularity-aware multi-domain universal image translation}
\label{subs:image_translation}
Buildings are artificial objects, and their fa\c{c}ades have regularly distributed components. While previous data-driven deep learning image synthesis or translation methods have made great progress on many datasets, these methods usually aim at non-artificial images, such as faces and natural scenes, which do not require clear boundaries \citep{karras2019style}. They have limitations on images with highly structured, man-made objects, including building fa\c{c}ades. Additionally, existing methods cannot preserve enough high-frequency details on synthesized results, which affects the expression of architectural style and realism \citep{cai2021frequency}. Therefore, to address these issues, we use frequency and corner information from the pixel and spectral domains to improve the synthesized results. Moreover, due to the different styles of buildings, this paper also utilizes a style encoder to specify the style and embeds the SEAN module \citep{zhu2020sean} in the generator to achieve the fa\c{c}ade texture translation of arbitrary style buildings from a semantic label. The objective of our network can be formally summarized by the following equations.
\begin{equation}
	\label{equ:opti}
	\underset{\boldsymbol{E},\boldsymbol{G}}{\min} \underset{\boldsymbol{D}}{\max}\Upsilon(\boldsymbol{E},\boldsymbol{D},\boldsymbol{G})
\end{equation}
where $\boldsymbol{E}, \boldsymbol{D}, \boldsymbol{G}$ are trainable style encoder, discriminator and generator respectively. $\Upsilon$ is the measurement of difference between ground-truth images and synthetic images. The meaning of $\Upsilon$ is shown in Equation \ref{equ:upsilon}.
\begin{equation}
	\label{equ:upsilon}
	\Upsilon(\boldsymbol{E},\boldsymbol{D},\boldsymbol{G}) = \mathcal{V}_{GAN} + \mathcal{V}_{detail} + \mathcal{V}_{regularity}
\end{equation}
$\Upsilon$ comprises of $\mathcal{V}_{GAN}$, $\mathcal{V}_{detail}$, and $\mathcal{V}_{regularity}$, which measure the dissimilarity between the synthesized image and the ground-truth image from different perspectives. $\mathcal{V}_{GAN}$ denotes the conventional GAN measurement, while $\mathcal{V}_{detail}$ and $\mathcal{V}_{regularity}$ enhance the synthetic results by enriching details and improving regularity, respectively. Their specific forms are shown in Equation \ref{equ:v}.
\begin{equation}
	\label{equ:v}
	\begin{cases}
		\mathcal{V}_{detail} = \mathcal{V}\left(\mathcal{F}\left( \mathbf{R'}\right), \mathcal{F}( \mathbf{R})\right)
		\\\mathcal{V}_{regularity}= \mathcal{V}\left(\mathcal{C}\left( \mathbf{R'}\right), \mathcal{C}( \mathbf{R})\right)
	\end{cases}
\end{equation}
where $\mathcal{V}$ denotes the difference between two samples; $\mathcal{F}(\cdot )$ and $\mathcal{C}(\cdot )$ are frequency transformation and corner map extraction respectively. $\mathcal{F}(\cdot )$ is calculated in multiple domains by Equation \ref{equ:high_frequency} (pixel domain) and \ref{equ:real} (spectral domain). Equations \ref{equ:sobel} to \ref{equ:R} are the specific calculation process of $\mathcal{C}(\cdot )$.
\subsubsection{Network architecture}
\begin{figure}[h]
	\centering
	\includegraphics[width=\linewidth]{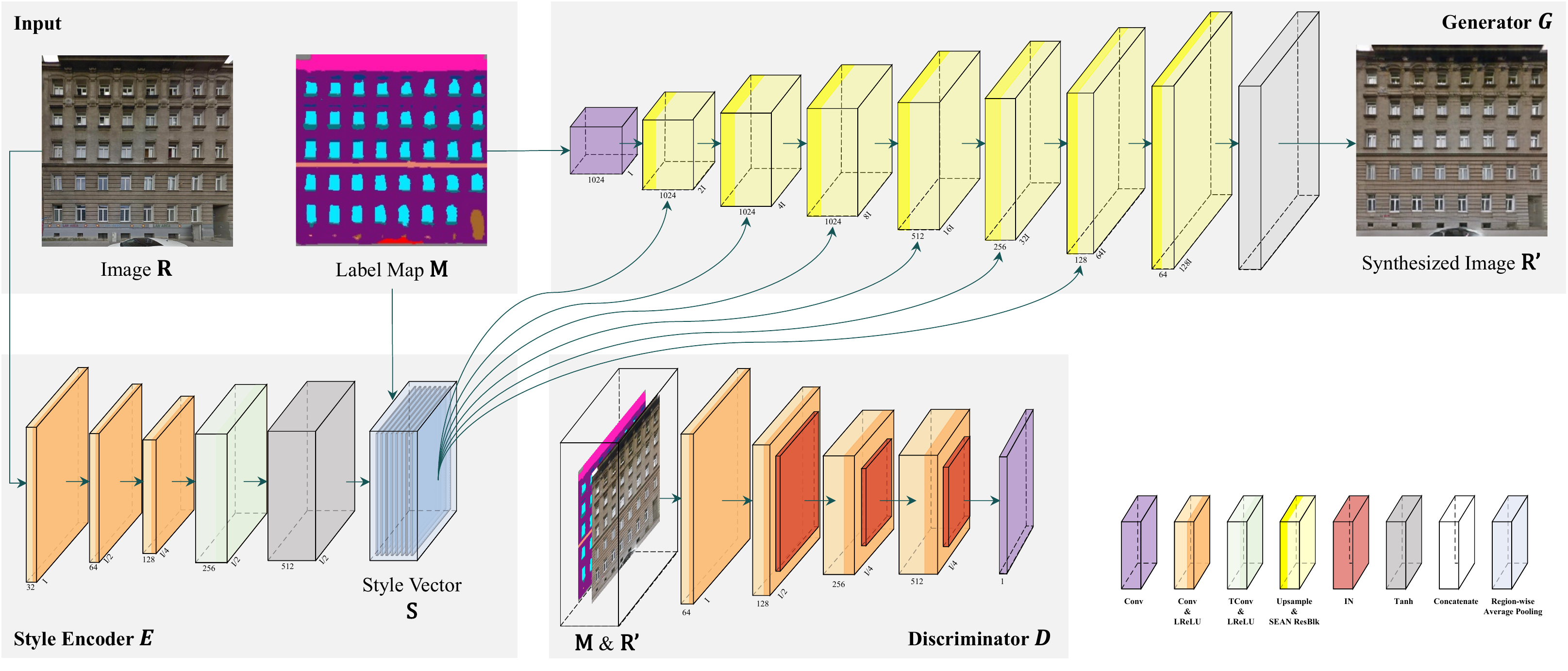}
	\caption{Network architecture.}
	\label{fig:net}
\end{figure}
Figure \ref{fig:net} depicts the architecture of our image translation network, which takes different inputs during training and inference phases. In the training phase, the inputs are a fa\c{c}ade image $\mathbf{R}$ with its corresponding semantic label map $\mathbf{M}$. In the inference phase, another semantic label map $\mathbf{M'}$ is required to provide the synthesis content. The network comprises a trainable style encoder $\boldsymbol{E}$, generator $\boldsymbol{G}$, and discriminator $\boldsymbol{D}$. $\boldsymbol{E}$ encodes the fa\c{c}ade image into a style vector $\mathbf{S}$ for every class based on the semantic label map. $\boldsymbol{G}$ synthesizes an image from the input label map $\mathbf{M'}$, which is equivalent to $\mathbf{M}$ during training. $\boldsymbol{D}$ is utilized to implicitly evaluate the synthesized result.
The structure of the generator and discriminator is shown in Figure \ref{fig:net}. The employed generator, with residual blocks, and the multi-scale discriminator follow the SPADE \citep{park2019semantic} and Pix2PixHD \citep{wang2018high} architectures, respectively. SPADE manipulates style transfer at the feature space, considering the statistical characteristics of the feature map as style and the normalized feature map as content \citep{park2019semantic}. Specifically, this approach first encodes the style image to a vector and then uses its statistical characteristics to de-normalize the input Gaussian-distributed vector of the generator, achieving state-of-the-art synthesized results with given styles. Our paper adopts this method to achieve universal image translation.

\begin{figure}[H]
	\centering
	\includegraphics[width=\linewidth]{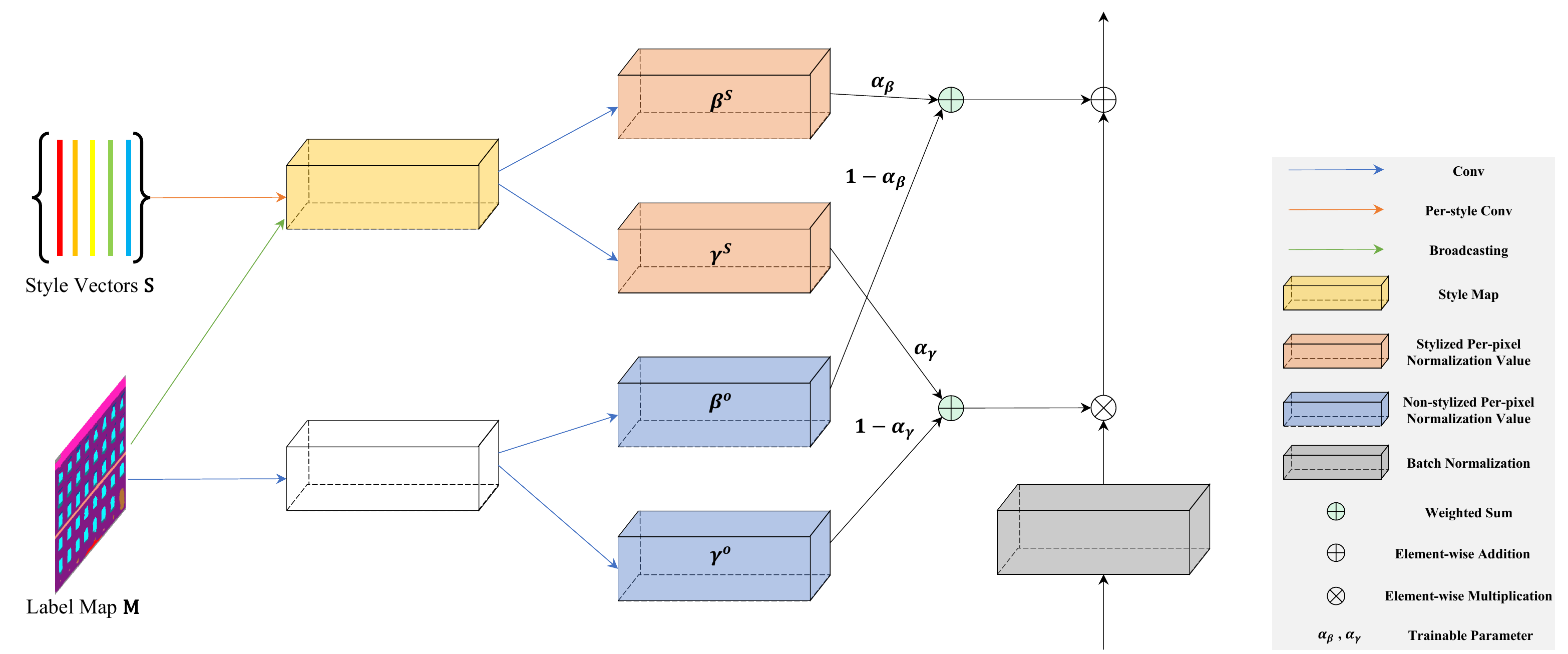}
	\caption{Structure of SEAN ResBlk.}
	\label{fig:sean}
\end{figure}

\paragraph{Semantic de-normalization}
As shown in Figure \ref{fig:net}, we employ the SEAN module \citep{zhu2020sean} which is an improved version of SPADE \citep{park2019semantic}. The structure of the SEAN module is shown in Figure \ref{fig:sean}. It embeds style and semantics in the generator through de-normalization operation. As depicted in Figure \ref{fig:sean}, the SEAN module consists of two parts. The upper part embeds image style into the network to generate images of the same style, while the lower part uses the SPADE structure to embed semantic information into the generation network to improve image generation quality \citep{park2019semantic}. Specifically, the activation value at position $(n \in N, c \in C, y \in H, x \in W)$ is calculated by Equation \ref{equ:act_value}:
\begin{equation}
	\label{equ:act_value}
	 \gamma_{c, y, x}(\mathbf{S}, \mathbf{M}) \frac{h_{n, c, y, x}-\mu_{c}}{\sigma_{c}}+\beta_{c, y, x} (\mathbf{S}, \mathbf{M})
\end{equation}
where $h$ is the activation value before normalization, $\mu$ and $\sigma$ are the mean and variance of the activation value on channel $c$, $\gamma$ and $\beta$ are calculated by Equation \ref{equ:gamma}:

\begin{equation}
	\label{equ:gamma}
	\begin{cases}
		\gamma_{c, y, x}(\mathbf{S}, \mathbf{M})=\alpha_{\gamma} \gamma_{c, y, x}^{s}(\mathbf{S})+\left(1-\alpha_{\gamma}\right) \gamma_{c, y, x}^{o}(\mathbf{M}) \\
		\beta_{c, y, x}(\mathbf{S}, \mathbf{M})=\alpha_{\beta} \beta_{c, y, x}^{s}(\mathbf{S})+\left(1-\alpha_{\beta}\right) \beta_{c, y, x}^{o}(\mathbf{M})
	\end{cases}
\end{equation}
where $\alpha_{\gamma}$ and $\alpha_{\beta}$ are the trainable parameters. $\gamma^s$, $\beta^s$ are obtained from convolution with style vector $\mathbf{S}$ as input, and $\gamma^o$, $\beta^o$ are obtained from semantic label map $\mathbf{M}$ after convolution.
\paragraph{Style encoder}
The encoder in Figure \ref{fig:net} takes an input image to extract deep features using three convolution layers with leaky rectified linear unit (LReLU) activation and down-sampling. A transpose convolution layer and a Tanh activation layer are then employed to reconstruct the image. The activated feature map is then region-wise average pooled according to the semantic label map corresponding to the input style image. The output is a set of vectors, each containing style information of its corresponding semantic category, which can be utilized for style transfer.
\paragraph{Generator}
The generator in Figure \ref{fig:net} employs the structure of the Pix2PixHD generator \citep{wang2018high} and replaces the residual module with the SEAN module to achieve style embedding with semantic information. A $3 \times 3$ convolution is used first, followed by seven SEAN ResBlks (shown in Figure \ref{fig:sean}), and a Tanh activation layer to obtain the output. Upsampling is performed before each SEAN ResBlk. The style vector extracted by the $\mathbf{E}$ encoder is injected into the first six SEAN ResBlks. The input and output of the generator $\mathbf{G}$ are the semantic label map $\mathbf{M}$ and the synthesized image $\mathbf{R}'$, respectively \citep{zhu2020sean}.
\paragraph{Discriminator}
To determine whether the high-resolution synthesized image is real or fake, a discriminator with a large receptive field is needed. However, using a larger convolution kernel and a deeper network will increase unnecessary computing costs and may lead to overfitting. Therefore, we employ a multi-scale discriminator with instance normalization (IN) and LReLU activation. The structure of each discriminator is shown in Figure \ref{fig:net}. Two discriminators are used, and GAN loss is calculated by referring to PatchGAN \citep{wang2018high}, which improves the discriminator by enlarging the receptive field without increasing the network parameters. The inputs of the discriminators are the concatenation of the image and its corresponding semantic label, and the output is an estimation of the true and fake probability of the generated samples.

\subsubsection{Multi-domain losses}
In order to solve the problem that image translation cannot retain the structural features of source domain, a frequency domain adaptation approach is proposed \citep{cai2021frequency}. This method can preserve the high-frequency details of the source domain for better results in several tasks. Although our paper combines image translation with style transfer, the lack of source domain structural features is still inevitable. Thus, we use the frequency domain adaptation method to calculate and extract the frequency map (gradient map) and spectrum map corresponding to the image, and retain the consistency between the synthesized result and the real image in both pixel space and Fourier spectral space to ensure that the generated result maintains more details.

Furthermore, we observe that fa\c{c}ade textures have obvious structural characteristics and artificial rules. The optimization of frequency domain adaptation can only enrich the texture details, and is powerless in terms of structural information. Therefore, our paper proposes a regular optimization method that utilizes corner information in the pixel domain to reflect the regular structure of building fa\c{c}ades. This regular optimization method allows us to achieve regular optimization of image synthesis in the training process.

\begin{figure}[H]
	\centering
	\includegraphics[width=\linewidth]{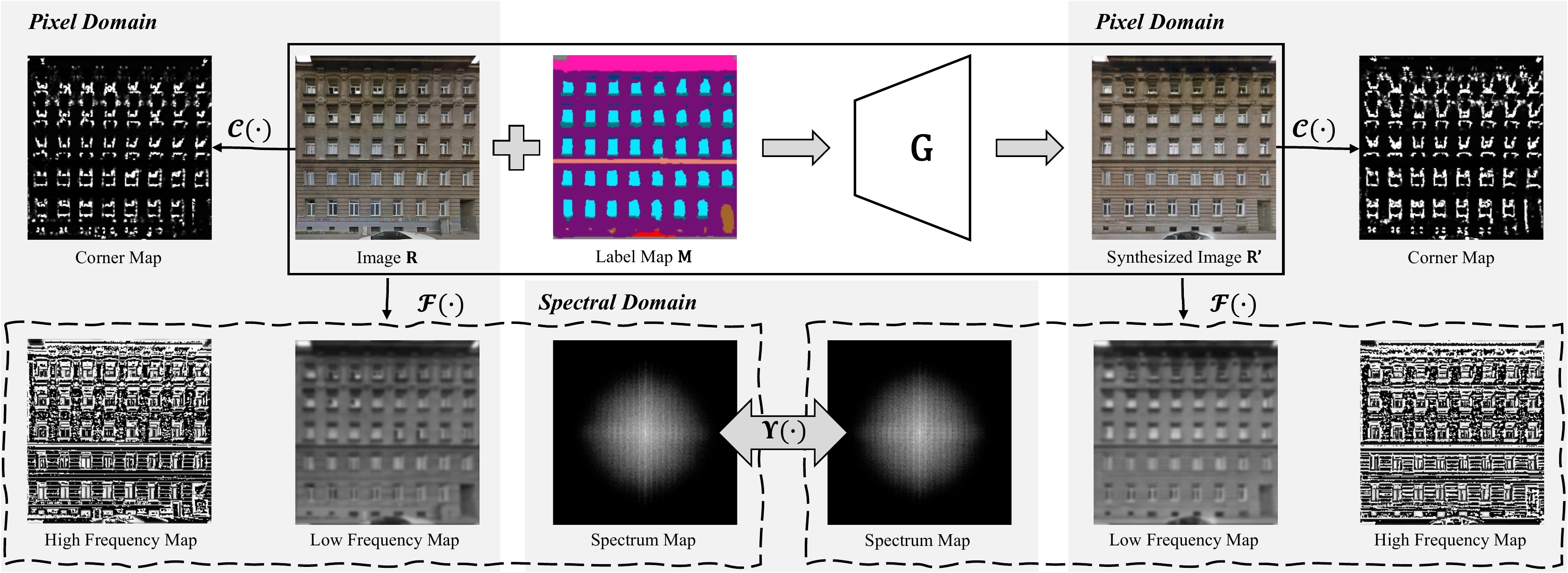}
	\caption{Multi-domain of proposed method. Pixel domain consists of original image and synthesized image, as well as their corresponding frequency maps and corner maps. Spectral domain is the spectrum maps of original image and synthesized image. The frequency maps and spectrum maps are obtained using the frequency domain adaptation method \citep{cai2021frequency}, while the corner maps are calculated using an improved Harris detector \citep{harris1988combined}.}
	\label{fig:domain}
\end{figure}

Overall, the multi-domain approach of our proposed method is illustrated in Figure \ref{fig:domain}. We measure and minimize the distance between synthesized images and input images in both pixel and spectral domains to optimize the network for better details and regularities in the results.

\paragraph{\textbf{{1)Pixel domain}}}
\paragraph{Frequency map}
We first perform low-pass filtering on the ground-truth image and synthetic image to obtain correspondence low-frequency information expressed in pixel domain. Then, the original images are converted into grayscale images and make difference with the low-frequency images to obtain the high-frequency information expressed in pixel domain. The network weights are optimized by comparing the difference of frequency information in pixel domain between the synthesized result and the ground-truth image, so that to retain more details in generated image.
More formally, this paper adopts Gaussian kernel to do low-pass filtering, the specific form is as follows:
\begin{equation}
	\label{equ:Gaussian}
	k_{\sigma}[i, j]=\frac{1}{2 \pi \sigma^{2}} e^{-\frac{1}{2}\left(\frac{i^{2}+j^{2}}{\sigma^{2}}\right)}
\end{equation}
where $[i, j]$ denotes the pixel location, $\sigma$ is standard deviation of Gaussian function. By using the Gaussian low-pass filter, we can get the frequency maps corresponding to the input image that can express high and low frequency information. The specific method is shown in following equation:
\begin{equation}
	\label{equ:high_frequency}
	\left\{\begin{array}{c}
		\mathrm{I}_{L}=k \otimes \mathrm{I} \\
		\mathrm{I}_{H}=\mathcal{G}(\mathrm{I})-\mathrm{I}_{L}
	\end{array}\right.
\end{equation}
where $\mathrm{I}$, $\mathrm{I}_{L}$ and $\mathrm{I}_{H}$ denote the image and its correspondence low and high frequency map respectively. $k$ is Gaussian kernel and $\otimes$ is convolution operation. $\mathcal{G}$ is the function that can convert an image from RGB color space to grayscale space.

\paragraph{Corner map}
\label{para:corner_loss}
Traditional corner detection methods usually first calculate the gradient map of the horizontal and vertical direction and then determine whether the pixel is a corner point by using a threshold value. However, this process is not differentiable, which can pose a serious problem for backpropagation optimization in deep convolutional networks. To address this issue, we propose an optimized Harris detector \citep{harris1988combined} that is differentiable and suitable for use in deep learning methods.

\begin{figure}[H]
\centering
\subcaptionbox{Fa\c{c}ade image}[0.48\linewidth]{
	\includegraphics[width=0.48\linewidth]{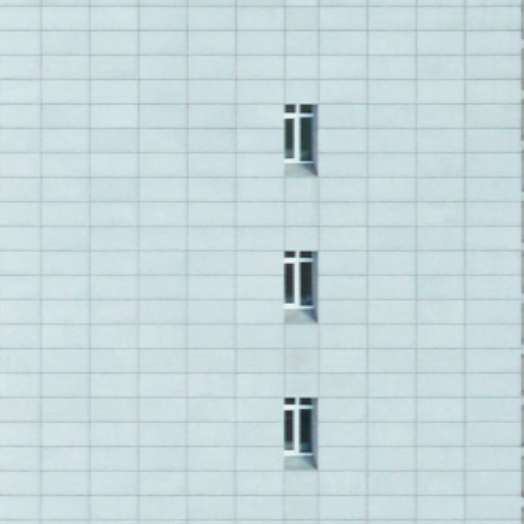}
	\includegraphics[width=0.48\linewidth]{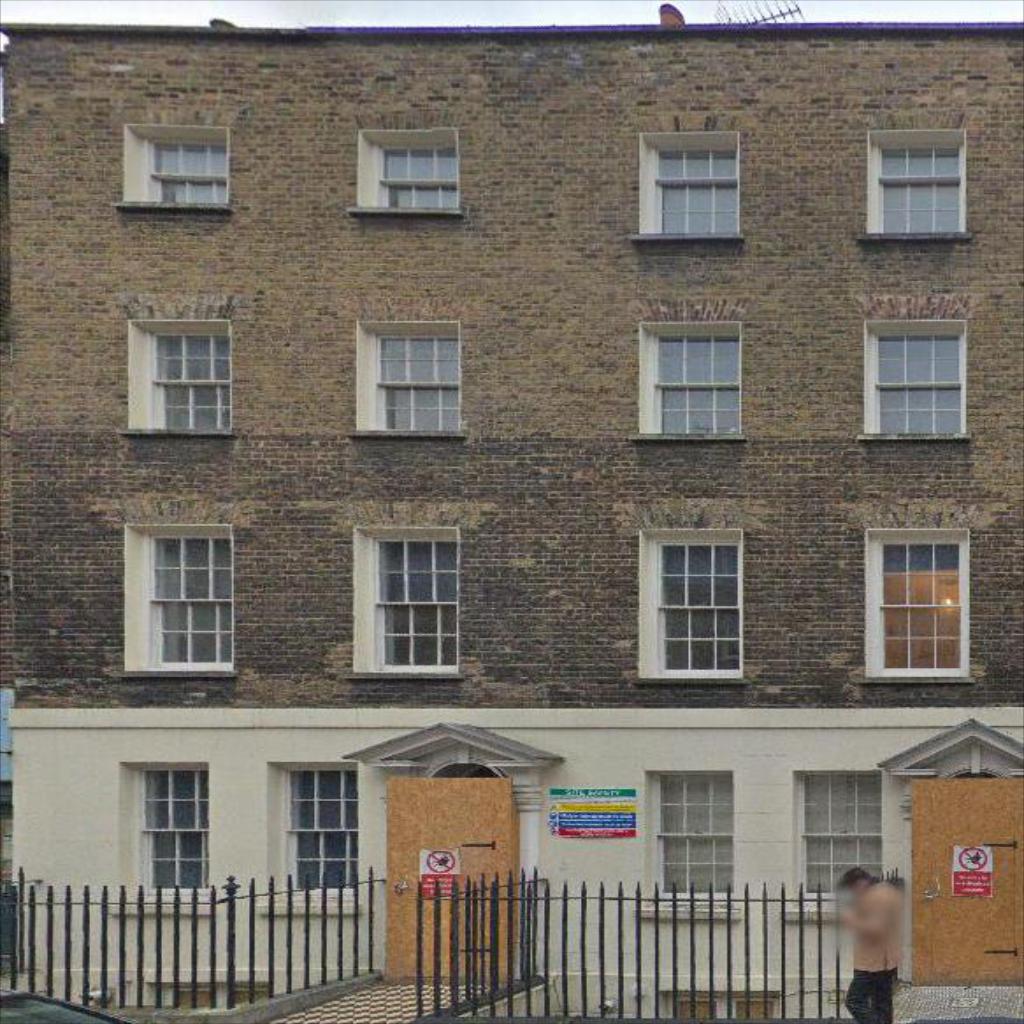}
}
\subcaptionbox{Visualized $R^{*}$}[0.48\linewidth]{
	\includegraphics[width=0.48\linewidth]{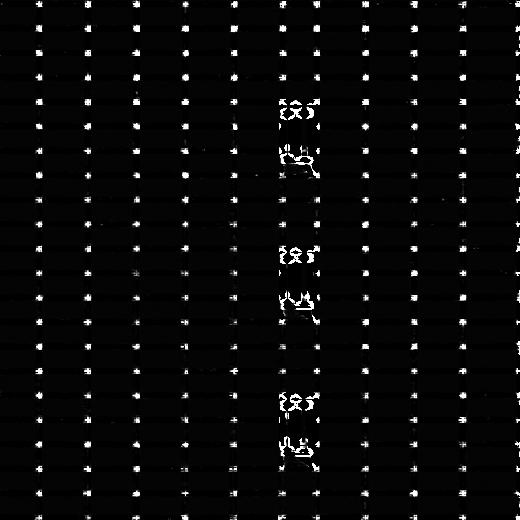}
	\includegraphics[width=0.48\linewidth]{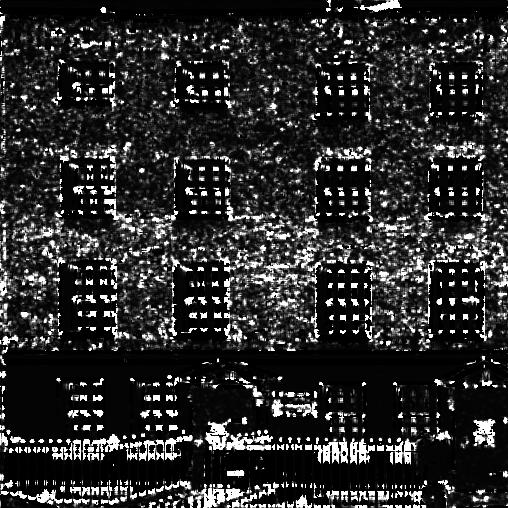}
}
\caption{Fa\c{c}ade images and their corresponding visualized $R^{*}$.}
\label{fig:corner}
\end{figure}

As shown in Equation \ref{equ:sobel}, we use Sobel operator $S_{x}$ and $S_{y}$ to extract gradient information in horizontal and vertical direction of the image $\mathrm{I}$.
\begin{equation}
	\label{equ:sobel}
	\left\{\begin{array}{l}
		\mathrm{I}_{x}=\mathrm{I} \otimes S_{x} \\
		\mathrm{I}_{y}=\mathrm{I} \otimes S_{y}
	\end{array}\right.
\end{equation}
where $\mathrm{I}_{x}$ and $\mathrm{I}_{y}$ are the gradient maps in two different directions. The product between the two directional gradients and their squares is then calculated, as shown in the following equation.
\begin{equation}
	\left\{\begin{array}{l}
		\mathrm{I}_{x}^{2}=\mathrm{I}_{x} \circ \mathrm{I}_{x} \\
		\mathrm{I}_{y}^{2}=\mathrm{I}_{y} \circ \mathrm{I}_{y} \\
		\mathrm{I}_{x} \mathrm{I}_{y}=\mathrm{I}_{x} \circ \mathrm{I}_{y}
	\end{array}\right.
\end{equation}
where $\circ$ denotes the pixel-wise product. After that, we calculate the Gaussian weighted sum for $\mathrm{I}_{x}^{2}$, $\mathrm{I}_{y}^{2}$ and $\mathrm{I}_{x} \mathrm{I}_{y}$.
\begin{equation}
	\label{equ:gaussian_sum}
	M=\sum_{(x, y) \in W} w_{G}(x, y)\left[\begin{array}{cc}
		\mathrm{I}_{x}^{2} & \mathrm{I}_{x} \mathrm{I}_{y} \\
		\mathrm{I}_{x} \mathrm{I}_{y} & \mathrm{I}_{y}^{2}
	\end{array}\right]
\end{equation}
where $w_{G}$ is the window function, which is set to Gaussian kernel function in this paper. $W$ denotes the current sliding window being processed. Subsquently, we can get corner response matrix $R$ by Equation \ref{equ:r}. 
\begin{equation}
	\label{equ:r}
	R = \det M - k (\mathrm{trace} M)
\end{equation}
where $\det$ and $\mathrm{trace}$ denote the determinant and trace of matrix $M$. According to \cite{harris1988combined}, pixels with $R$ values greater than, equal to, and less than zero are considered corner points, flat areas, and edges, respectively. In this paper, we use a differentiable rectified linear unit (ReLU) function to remove negative and zero values that are not relevant, and obtain an equivalent expression of corner information through scaling. The specific form is shown in Equation \ref{equ:R}.
\begin{equation}
	\label{equ:R}
	R^{*}=\omega \cdot \mathrm{ReLU}(R)
\end{equation}
where $\omega = 100000$ is the scaling factor, and as shown in Figure \ref{fig:corner}, $R^{*}$ is the final value matrix that represents the corner information in the pixel domain.
\paragraph{\textbf{{2)Spectral domain}}}
\paragraph{Spectrum map}
The pixel domain information is presented in the form of coordinates, and global information cannot be considered. The calculation of each position in the Fourier spatial spectrum map needs to use the information of all pixels, so it can reflect the global features of the image. Therefore, in addition to pixel domain optimization, this paper also uses Fast Fourier Transform (FFT) to convert image $\mathrm{I}$ from pixel domain to Fourier spectral domain. The neural network weights are optimized by comparing the spectrum difference between the synthesized result and ground-truth fa\c{c}ade image, thus further retaining the consistency globally. The Discrete Fourier Transform (DFT) applied to a single image $\mathrm{I}$ can be formally expressed as follows:
\begin{equation}
	\label{equ:fft}
	\mathcal{F}(\mathrm{I})(x, y)=\frac{1}{H W} \sum_{h=0}^{H-1} \sum_{w=0}^{W-1} e^{-2 \pi i \cdot \frac{h x}{H}} e^{-2 \pi i \cdot \frac{w y}{W}} \cdot \mathrm{I}(h, w)
\end{equation}
where $(x, y)$ denotes the pixel location in image $\mathrm{I}\in\mathbb{R}^{H \times W }$. Since the values of $\mathcal{F}(\mathrm{I})$ are complex numbers, we use Equation \ref{equ:real} to transform them into a field of real numbers.
\begin{equation}
	\label{equ:real}
	\mathcal{F}^{R}(\mathrm{I})(x, y)=\log \left(1+\sqrt{\left[\mathcal{F}_{R}(\mathrm{I})(x, y)\right]^{2}}\right. \\
	\left.+\sqrt{\left[\mathcal{F}_{I}(\mathrm{I})(x, y)\right]^{2}}+\epsilon\right)
\end{equation}
where $\mathcal{F}_{R}(\mathrm{I})$ and $\mathcal{F}_{I}(\mathrm{I})$ denote the real and imaginary parts of $\mathcal{F}(\mathrm{I})$. $\epsilon$ is an additional term of numerical stability, this paper is set to $1 \times 10^{-8}$ \citep{cai2021frequency}. $\mathcal{F}^{R}(\mathrm{I})$ is the final spectrum map of $\mathrm{I}$ to be compared during training.
\paragraph{\textbf{{3)Overall loss}}}
The overall loss function can be formally expressed by following equation:
\begin{equation}
	\label{equ:losses}
	\min _{\mathbf{E}, \mathbf{G}}\left(\left(\max _{\mathbf{D_{1}}, \mathbf{D_{2}}} \sum_{k=1,2} \mathcal{L}_{GAN}\right)+\lambda_{1} \sum_{k=1,2} \mathcal{L}_{FM}+\lambda_{2} \mathcal{L}_{P} +\lambda_{3} \mathcal{L}_{D}+\lambda_{4} \mathcal{L}_{\text {R}}\right)
\end{equation}
where the five terms represent GAN loss, feature matching loss, perceptual loss, detail loss and regularity loss respectively. These terms are introduced in detail in the following paragraphs. $\mathbf{D_{1}}$ and $\mathbf{D_{2}}$ denote two PatchGAN discriminators. The weights to balance the different losses are set to $\lambda_{1}=\lambda_{2}=\lambda_{3}=\lambda_{4}=10$ and $\lambda_{5}=5\times 10^{-6}$.
\paragraph{GAN loss}
The form of the GAN loss function in this paper is shown in Equation \ref{equ:gan_loss}. The optimization is to minimize the loss of style encoder $\mathbf{E}$ and generator $\mathbf{G}$, and maximize the loss of discriminators $\mathbf{D_1}, \mathbf{D_2}$. The main objective is to train generator $\mathbf{G}$ to generate high-quality images that can deceive discriminator $\mathbf{D}$, making the discriminator unable to distinguish between real and synthesized images.
\begin{equation}
	\label{equ:gan_loss}
	\min _{\mathbf{E}, \mathbf{G}} \max _{\mathbf{D_{1}}, \mathbf{D_{2}}} \sum_{k=1,2} \mathcal{L}_{G A N}\left(\mathbf{E}, \mathbf{G}, \mathbf{D_{k}}\right)
\end{equation}
The Hinge loss function used for $\mathcal{L}_{GAN}$ in this paper is defined as follows:
\begin{equation}
	\label{equ:hinge}
		\mathcal{L}_{GAN}=\mathbb{E}[\max ( \left.\left.0,1-\mathbf{D_{k}}(\mathbf{R}, \mathbf{M})\right)\right]
		+\mathbb{E}[\max (0,1\left.\left.-\mathbf{D_{k}}(\mathbf{G}(\mathbf{S}, \mathbf{M}), \mathbf{M})\right)\right]
\end{equation}
Refer to section \ref{subsubs:problem_setup} for the meaning of parameters in the formula.
\paragraph{Feature matching loss}
The feature matching loss evaluates the difference between the feature maps of the ground-truth image and the synthetic image at different levels, which are extracted from the discriminator network. Optimizing the network weights according to the feature matching loss can achieve better consistency at the feature level, resulting in better results. The definition of the feature matching loss is shown as follows:
\begin{equation}
	\label{equ:fm}
	\mathcal{L}_{F M}=\mathbb{E} \sum_{i=1}^{T} \frac{1}{N_{i}}  {\left[\| \mathbf{D_{k}^{(i)}}(\mathbf{R}, \mathbf{M})\right.}\left.-\mathbf{D_{k}^{(i)}}(\mathbf{G}(\mathbf{S}, \mathbf{M}), \mathbf{M}) \|_{1}\right]
\end{equation}
where $D_{k}^{(i)}$ and $N_{i}$ denote the feature map outputted from the i-th layer of the k-th discriminator and the number of corresponding elements. $T$ is the total layer numbers of discriminator \citep{mirza2014conditional}.
\paragraph{Perceptual loss}
Perceptual loss measures the difference between two images by comparing the features extracted from the network pre-trained on large datasets, e.g. pre-trained VGG-16 is used in this paper. Specifically, perceptual loss can be calculated using the following equation:
\begin{equation}
	\label{equ:p_loss}
	\mathcal{L}_{P}=\mathbb{E} \sum_{i=1}^{N} \frac{1}{M_{i}}\left[\left\|F^{(i)}(\mathbf{R})-F^{(i)}(\mathbf{G}(\mathbf{S}, \mathbf{M}))\right\|_{1}\right]
\end{equation}
where $N$ is the layer number of pre-trained network, $F^{(i)}$ and $M_{i}$ denote the feature map extracted from i-th layer of pre-trained network and the number of corresponding elements \citep{johnson2016perceptual}.
\paragraph{Detail loss}
Detail loss is shown in Equation \ref{equ:E}, and consists of two parts, i.e., $\mathcal{L}_{D_{pix}}$ and $\mathcal{L}_{D_{spectral}}$.
\begin{equation}
	\label{equ:f_loss}
	\mathcal{L}_{D}=\mathcal{L}_{D_{pix}}+\mathcal{L}_{D_{spectral}}
\end{equation}
where $\mathcal{L}_{D_{pix}}$ and $\mathcal{L}_{D_{spectral}}$ denote the different detail losses, they come from frequency information in pixel and spectral domain respectively, they are calculated by following equations:
\begin{equation}
	\label{equ:pix}
		\mathcal{L}_{D_{pix}}=  \mathbb{E}\left[\left\|\mathbf{R}_{L}-\left(\mathbf{G}(\mathbf{S}, \mathbf{M})_{L}\right)\right\|_{1}\right. 
		 \left.+\left\|\mathbf{R}_{H}-\left(\mathbf{G}(\mathbf{S}, \mathbf{M})_{H}\right)\right\|_{1}\right]
\end{equation}
where $\mathbf{R}_{L}$ and $\mathbf{R}_{H}$ denote the low and high frequency map of ground-truth image, $\mathbf{G}(\mathbf{S}, \mathbf{M})_{L}$ and $\mathbf{G}(\mathbf{S}, \mathbf{M})_{H}$ are the low and high frequency map of generated image. Frequency map is obtained from Equation \ref{equ:high_frequency}.
\begin{equation}
	\label{equ:fft_loss}
	\mathcal{L}_{D_{spectral}}=\mathbb{E}\left[\left\|\mathcal{F}^{R}(\mathbf{R})-\mathcal{F}^{R}(\mathbf{G}(\mathbf{S}, \mathbf{M}))\right\|_{1}\right]
\end{equation}
where $\mathcal{F}^{R}$ refers to the FFT and real number operation using Equation \ref{equ:fft} and \ref{equ:real}.
\paragraph{Regularity loss}
 The regularity loss is obtained by calculating the difference of corner maps between the generated image and the ground-truth image. The specific form is shown in Equation \ref{equ:regularity_loss}.
 \begin{equation}
 	\label{equ:regularity_loss}
 	\mathcal{L}_{\text {R }}=\mathbb{E}\left[\|\mathcal{C}(\mathbf{R})-\mathcal{C}(\mathbf{G}(\mathbf{S}, \mathbf{M}))\|_{1}\right]
 \end{equation}
where $\mathcal{C}(\cdot)$ is the corner map calculation introduced in paragraph \ref{para:corner_loss}.
\subsection{Image quilting}
\label{subs:image_quilting}
Because deep learning-based image synthesis methods rely on massive training data, they may not be able to generalize well to data that has not been seen. While embedding image styles can alleviate this problem, there is still a significant gap between the generated results and actual images, particularly for highly structured repetitive textures. Therefore, when the generation ability of deep learning methods is insufficient, this paper applies a sample-based image quilting algorithm to the wall region to compensate for this limitation. This process can be expressed as follows:
\begin{equation}
	\label{equ:image_quilting}
	\begin{cases}
		\mathbf{R'} = \mathbf{R'} & \mathcal{V}( \mathbf{R'},  \mathbf{R}) \le threshold
		\\\mathbf{R'} = \mathbf{R'}\otimes \neg  \mathbf{m'} \oplus \mathcal{Q}(\mathbf{R'}) \otimes  \mathbf{m'} &\mathcal{V}( \mathbf{R'},  \mathbf{R}) > threshold
	\end{cases}
\end{equation}
$\mathbf{R}'$ denotes the synthetic image of generator $\boldsymbol{G}$  on the right side of the equation, and the final result on the left side of the equation. $\mathbf{m'} \in\mathbb{B}^{H \times W}$ is a subset of $\mathbf{M'}$ which specifies the wall region in synthetic image. $\mathcal{Q}$ denotes image quilting which generates repetitive textures by stitching together small patches of un-occluded regions \citep{efros2001image}. Next, we will provide a detailed introduction to the image quilting algorithm $\mathcal{Q}$.

We begin the image quilting process by selecting a source pixel patch $\mathcal{S}$ from the synthetic image $\mathbf{R}'$. During image quilting, all pixels are sampled from $\mathcal{S}$. Firstly, we sample all $N \times N$ pixel patches $\mathcal{S}(\boldsymbol{p})$ from $\mathcal{S}$ to form the set $\mathcal{S}_B$. Next, we randomly select a patch $\mathcal{R}(\mathcal{S}_B)$ and define this patch as $B_1$. Subsequently, the algorithm iterates by a sequence $Q$ which contains the center position $\boldsymbol{p}$ of all patches in raster scanning order. The moving step during scanning is $\boldsymbol{v}$, which denotes the overlapping width. In this paper, the offset $\boldsymbol{v}$ is set to $(5, 0)$ in the first row, $(0, 5)$ in the first column, and $(5, 5)$ for the rest of the positions.
During iteration, we first find pixel patch $B_2$ from $\mathcal{S}_B$ by measuring its similarity with $B_1$. The overlapping regions in $B_1$ and $B_2$ are represented by $B_{ov}^1$ and $B_{ov}^2$ respectively. Next, we find a boundary $\boldsymbol{l}$ around the center line of the overlapping region by using the Dijkstra algorithm $\mathcal{D}$, based on the error $\mathcal{E}(B_{ov}^1- B_{ov}^2) ^2<\boldsymbol{t}$, where $\boldsymbol{t}=0.1$ denotes the minimum tolerance. After that, we update the pixel patches $\mathcal{T}(\boldsymbol{p})$ and $\mathcal{T}(\boldsymbol{p}+\boldsymbol{v})$ by stitching $B_1$ and $B_2$ using the minimum error boundary $\boldsymbol{l}$.
After the iteration process, we obtain the texture synthesized result $\mathcal{T}$ of image quilting.

\begin{algorithm}[H]
	\caption{Image quilting algorithm.}
	\label{algo:image_quilting}
	\begin{algorithmic}[2]
		\Procedure{ImageQuilting}{$\mathcal{S},\mathcal{T},Q,\boldsymbol{t},\boldsymbol{v}$}
		\State $\mathcal{S}_B\gets \left \{\mathcal{S}(\boldsymbol{p}) , \forall \boldsymbol{p} \in \mathcal{S} \right \}$ \Comment{Random Initialization}
		\State $B_1 \gets \mathcal{R}(\mathcal{S}_B)$
		
		\For{$\boldsymbol{p} \in Q$} \Comment{Raster Scanning}
			\State $B_2 \gets \mathcal{V}(B_1, \mathcal{S}_B)$
			\State $B_{ov}^1, B_{ov}^2 \gets B_1 \wedge B_2$
			\While{$\mathcal{E}(B_{ov}^1- B_{ov}^2) ^2<\boldsymbol{t}$} \Comment{Minimum Error Boundary Cut}
				\State $\boldsymbol{l} \gets \mathcal{D}(B_{ov}^1, B_{ov}^2)$
			\EndWhile
			\State $\mathcal{T}(\boldsymbol{p}), \mathcal{T}(\boldsymbol{p}+\boldsymbol{v}) \gets \mathcal{Q}(B_1, B_2, \boldsymbol{l})$ \Comment{Pixel Patch Quilting}
			\State $B_1 \gets B_2$
		\EndFor
		
		\EndProcedure
	\end{algorithmic}
\end{algorithm}

\section{Experimental evaluation and analysis}
\label{s:results}
\subsection{Dataset description}
\begin{figure}[H]
	\centering
	\subcaptionbox{London dataset}[0.48\linewidth]{
		\includegraphics[width=1\linewidth]{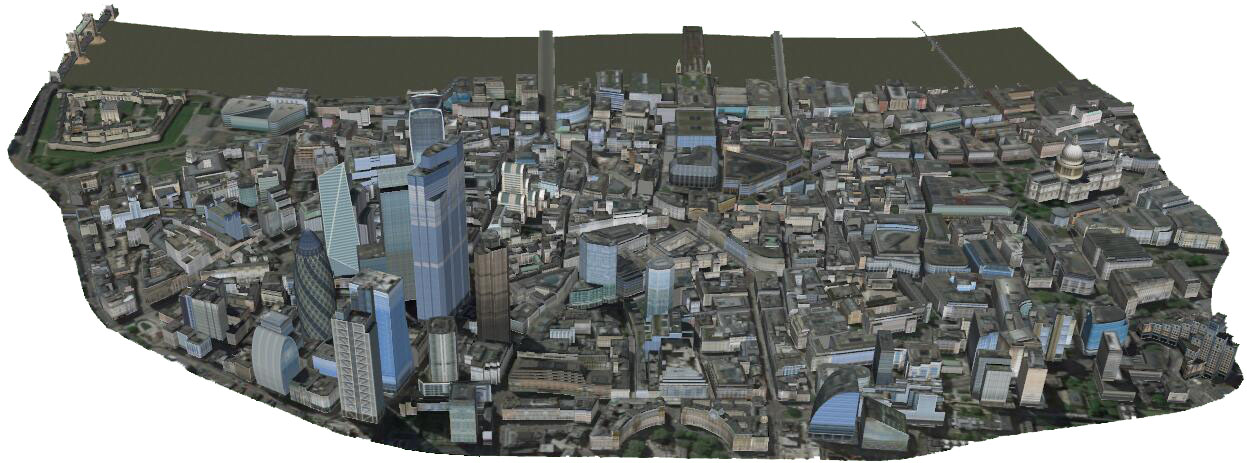}
	}
	\subcaptionbox{Wellington dataset}[0.48\linewidth]{
	\includegraphics[width=1\linewidth]{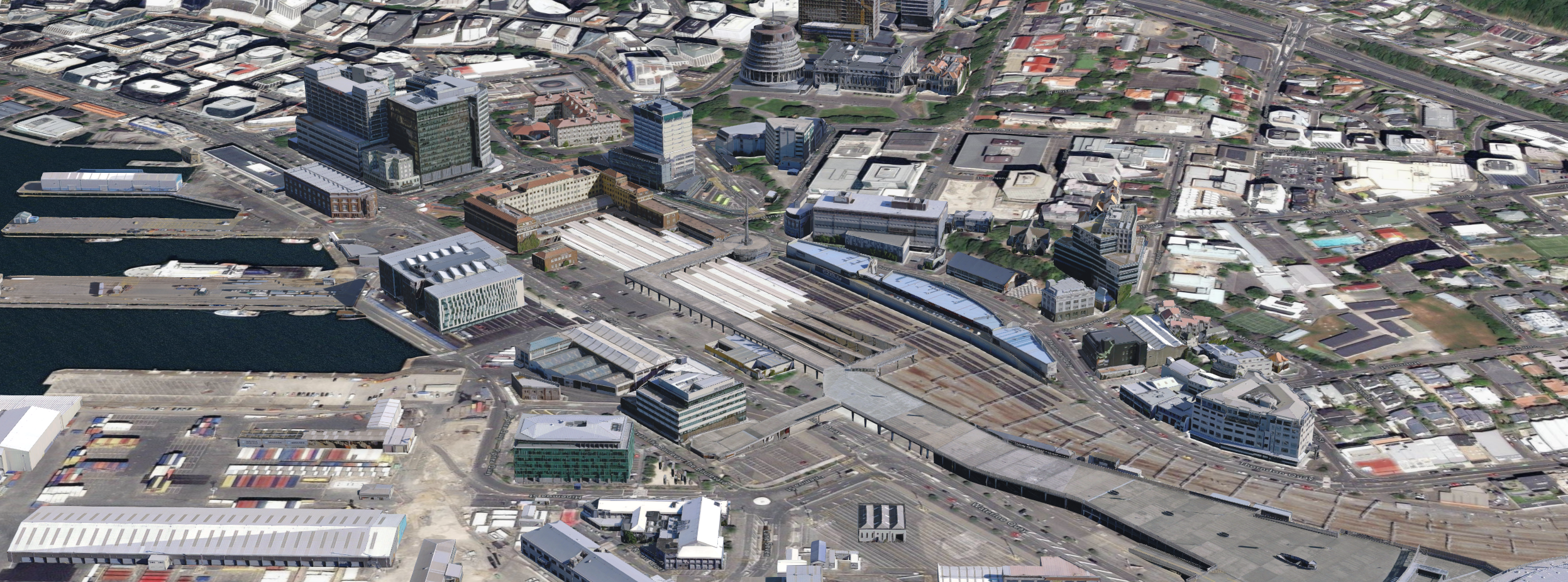}
}
	\caption{Datasets for experiment. London dataset is rendered in SketchUp with its own base map, and Wellington dataset is rendered in ArcGIS with public world imagery base map.They both are skp format which is modeled by SketchUp. The textures and base map of London dataset are the manual production of model creator. Due to the mutual occlusions between buildings, some building fa\c{c}ades have no texture. Unlike London dataset, the textures of Wellington dataset come from realistic scene, that contain several inevitable occlusions.}
	\label{fig:datasets}
\end{figure}

To verify the effectiveness of the proposed approach in this paper, we used two public urban textured 3D building datasets in .skp format from different countries as shown in Figure \ref{fig:datasets}. The first dataset was collected from central London, including several landmarks such as London Bridge. We selected some buildings with missing texture in this dataset for the experiment. The second dataset is from Pipitea South of Wellington, New Zealand, and we selected a typical block (Wellington train station) consisting of a series of buildings with occluded fa\c{c}ade textures to evaluate our approach. Moreover, we trained our proposed network using the Large Scale Architectural Asset (LSAA) dataset for performance evaluation. The LSAA dataset contains 199,723 fa\c{c}ade images of different styles extracted from large-scale rectified panoramas \citep{9145640}.
\subsection{Results}
\subsubsection{Semantics completion of building fa\c{c}ade}
\begin{figure}[H]
	\centering
	\subcaptionbox{Ground-truth}[0.23\linewidth]{
		\includegraphics[width=\linewidth]{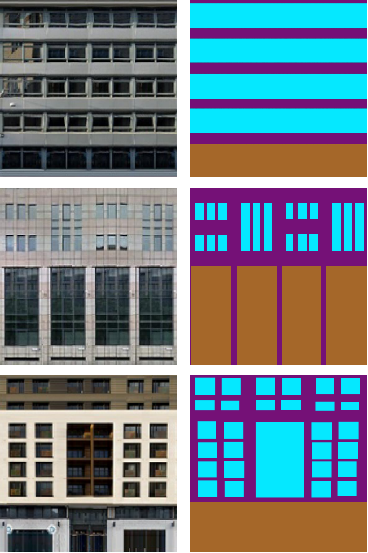}
	}
	\hspace{5mm}
	\subcaptionbox{Incomplete semantic map \& semantics completion results}[0.72\linewidth]{
		\includegraphics[width=\linewidth]{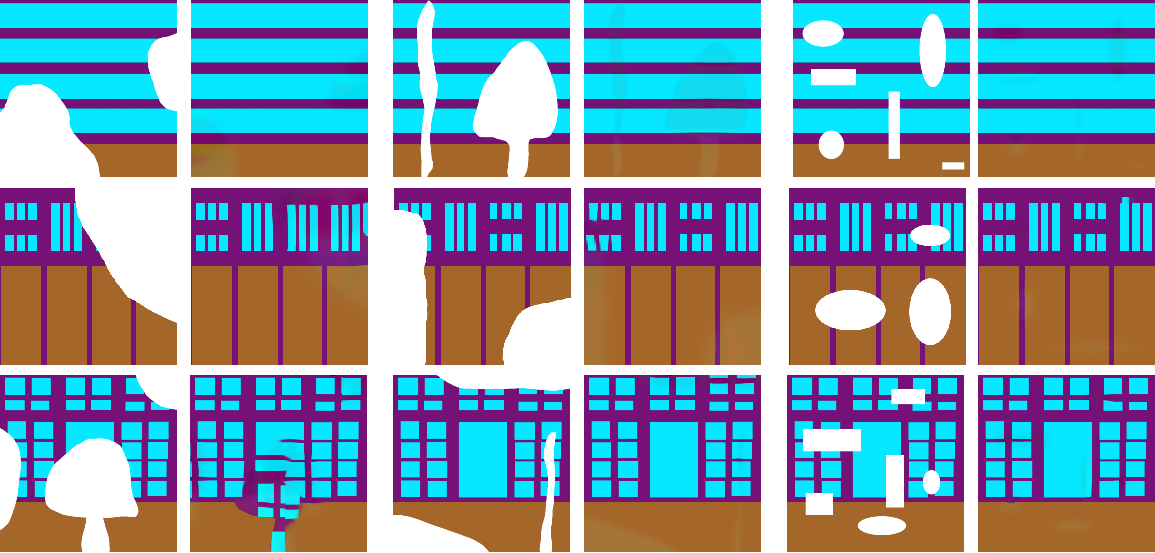}
	}	
	\caption{Hypothetical experiment of proposed semantics completion method. (a) Ground-truth non-occluded fa\c{c}de images and their correspondence semantic label maps. (b) The hypothetical incomplete images and semantics completion results recovered by proposed method. Different colors in (b) denote different semantics, i.e. window, wall and door.}
	\label{fig:semantice_completion_manual}
\end{figure}

\begin{figure}[H]
	\centering
	\subcaptionbox{Building fa\c{c}ade of Wellington train station}[0.47\linewidth]{
		\includegraphics[width=\linewidth]{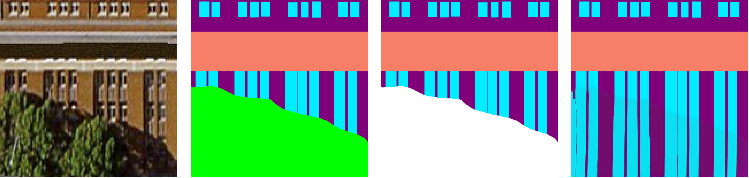}
	}
	\hspace{5mm}
	\subcaptionbox{Building fa\c{c}ade in Wellington}[0.47\linewidth]{
	\includegraphics[width=\linewidth]{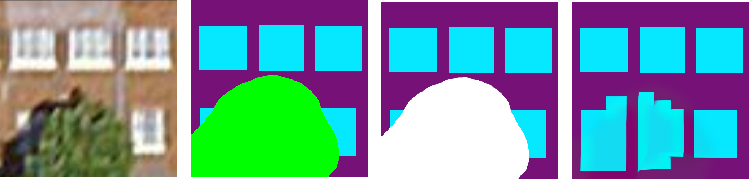}
}
	\subcaptionbox{Building fa\c{c}ade in London}[0.47\linewidth]{
	\includegraphics[width=\linewidth]{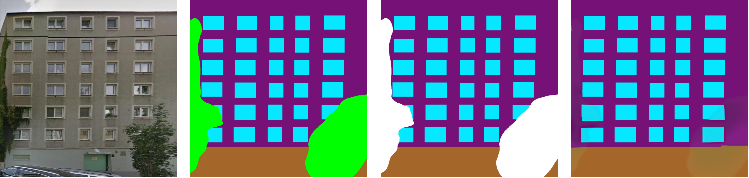}
}
	\hspace{5mm}
	\subcaptionbox{Building fa\c{c}ade in Australia}[0.47\linewidth]{
	\includegraphics[width=\linewidth]{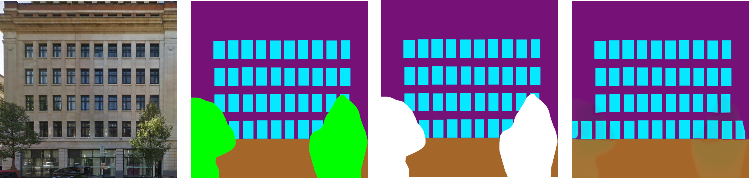}
}

	\caption{Semantics completion experiment on building fa\c{c}ades with occluded realistic textures. In each sub-figure, the occlusion texture and its corresponding semantic label image, mask image and recovery result are displayed from left to right. Different colors denote different semantics in label map, including window, wall, door, vegetation and cornice.}
	\label{fig:semantice_completion_occlusion}
\end{figure}

To solve the problem caused by frequent occlusions of building fa\c{c}ade in built-up areas, we first recover the semantics of occluded regions. Figure \ref{fig:semantice_completion_manual} and \ref{fig:semantice_completion_occlusion} show two different experiments to validate our semantics completion method. As shown in Figure \ref{fig:semantice_completion_manual}, we first consider non-occluded fa\c{c}ade images and their correspondence semantics as reference, then manually or randomly remove some regions of semantic label maps, finally adopt proposed method to recover the missing semantics. It can be seen that, except for the first group of experiments in the third row, the other experiments have obtained desirable results compared to the reference.

Figure \ref{fig:semantice_completion_occlusion} shows experiments on building fa\c{c}ades with occluded realistic textures, and reasonable results are obtained except for Figure \ref{fig:semantice_completion_occlusion}(b). By combining the experiments shown in Figure \ref{fig:semantice_completion_manual} and \ref{fig:semantice_completion_occlusion}, the proposed method demonstrates good performance on small occlusions. However, when the occlusion area is too large, achieving satisfactory recovery results may be challenging. Fortunately, by using the approach proposed in this paper, people can also manually annotate the semantics of building fa\c{c}ades for final texture synthesis. Moreover, the colors of different semantics are predetermined, so obtaining the label map with standard colors through simple correction after image completion is possible.

\subsubsection{Fa\c{c}ade texture repair of 3D building}
\begin{figure}[H]
	\centering
	\subcaptionbox{Unrepaired Wellington model}[0.47\linewidth]{
		\includegraphics[width=\linewidth]{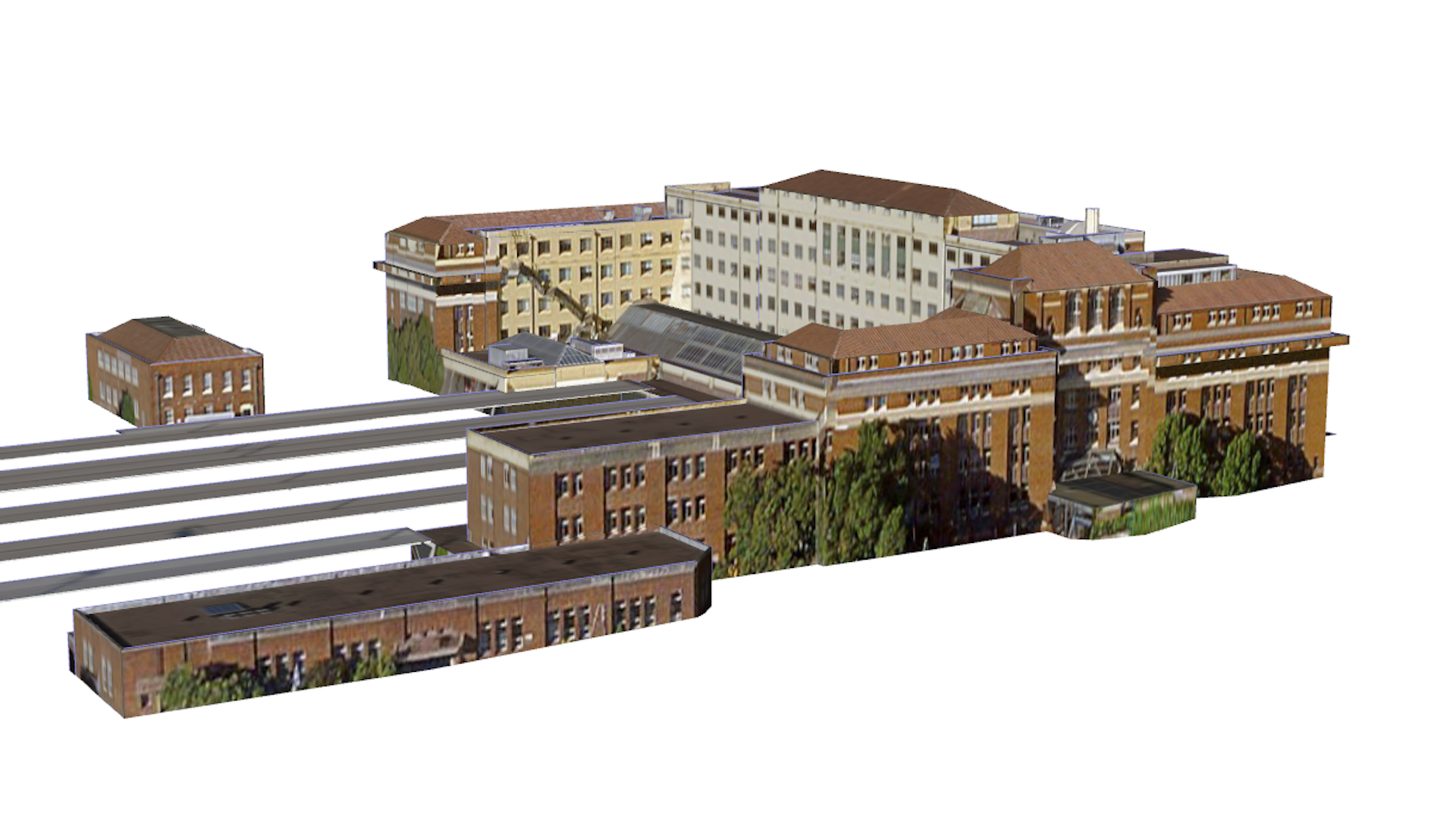}
	}
	\hspace{5mm}
	\subcaptionbox{Repaired Wellington model}[0.47\linewidth]{
		\includegraphics[width=\linewidth]{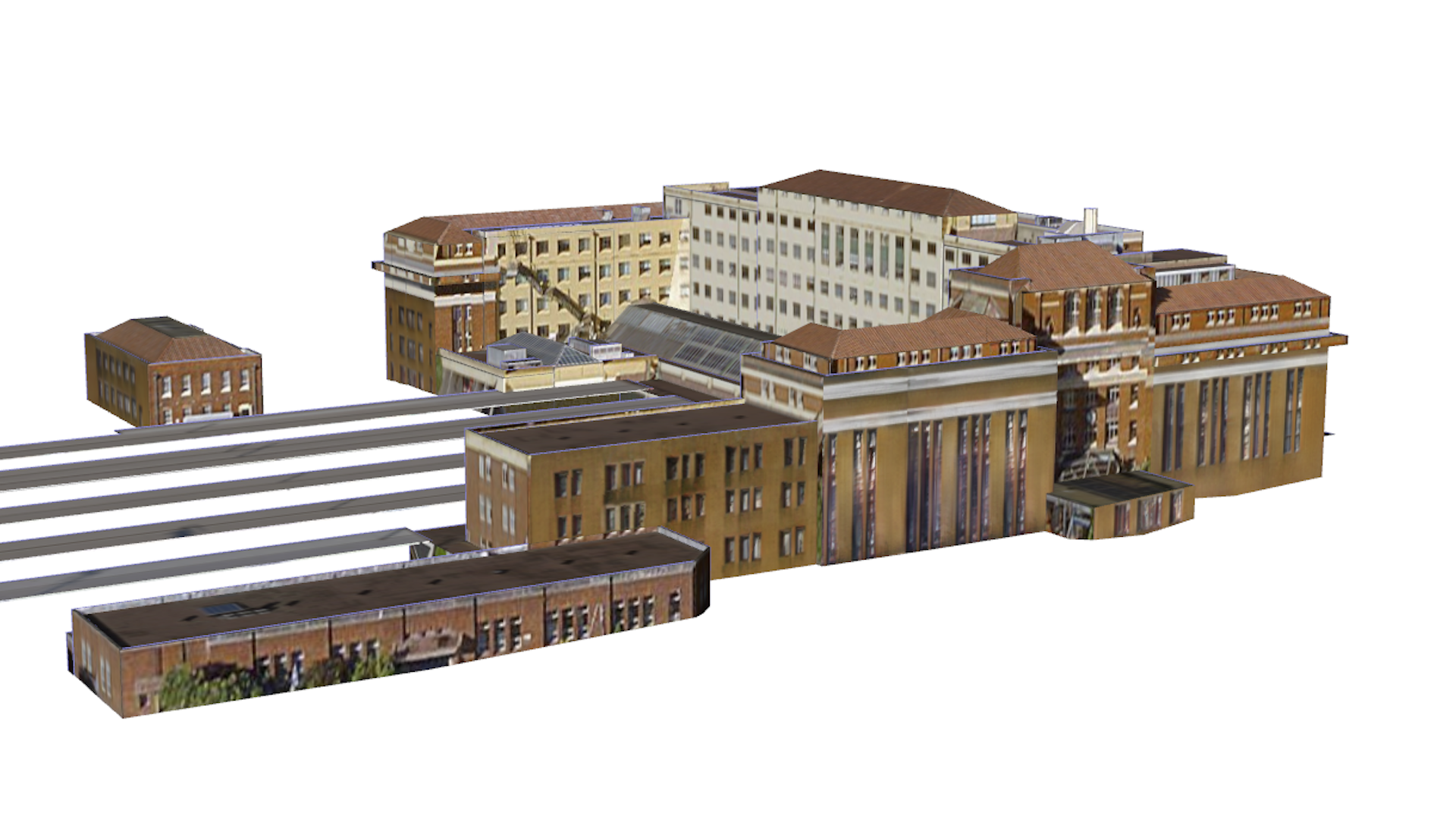}
	}
	
	\subcaptionbox{Unrepaired London model}[0.47\linewidth]{
		\includegraphics[width=\linewidth]{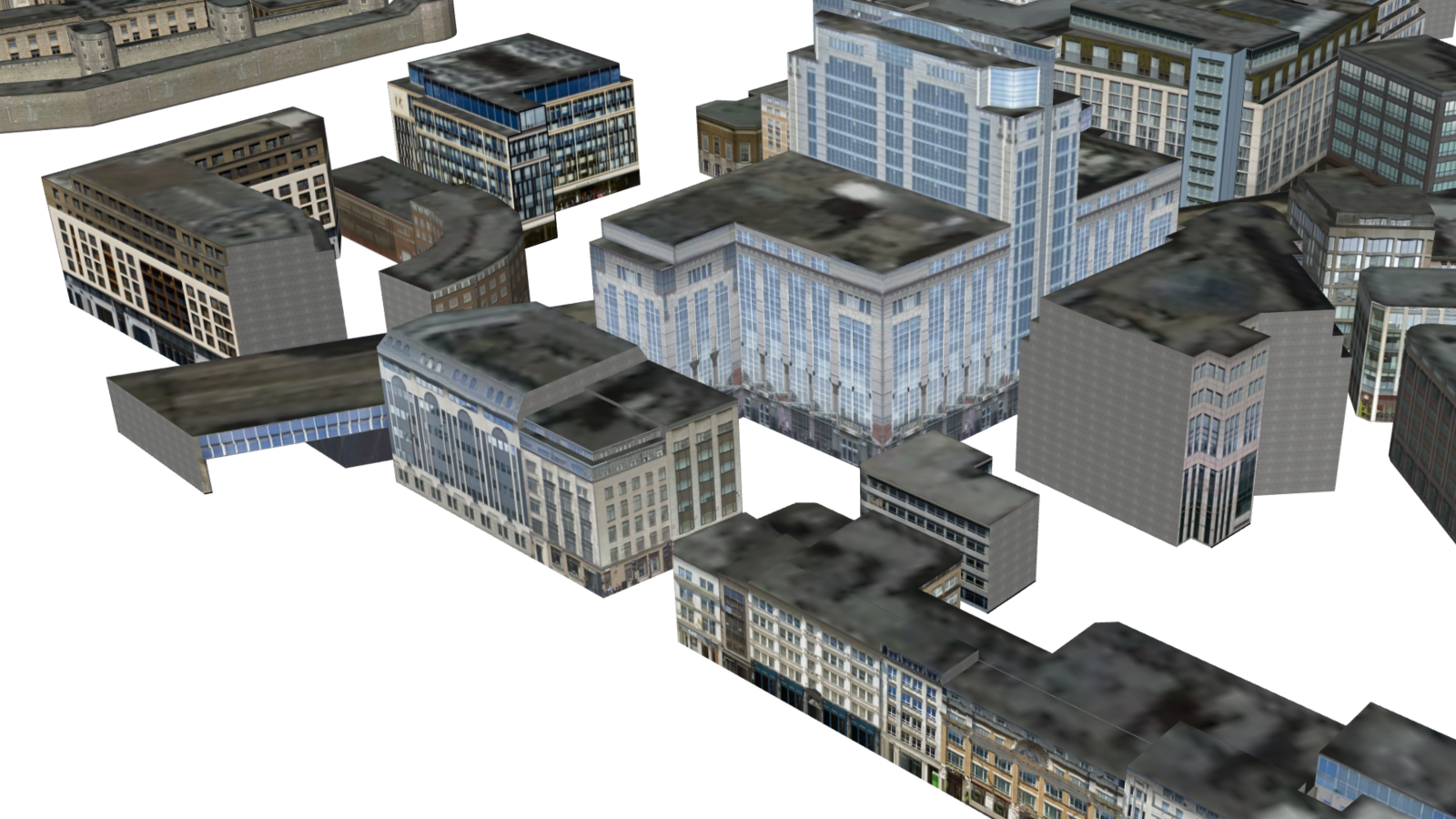}
	}
	\hspace{5mm}
	\subcaptionbox{Repaired London model}[0.47\linewidth]{
		\includegraphics[width=\linewidth]{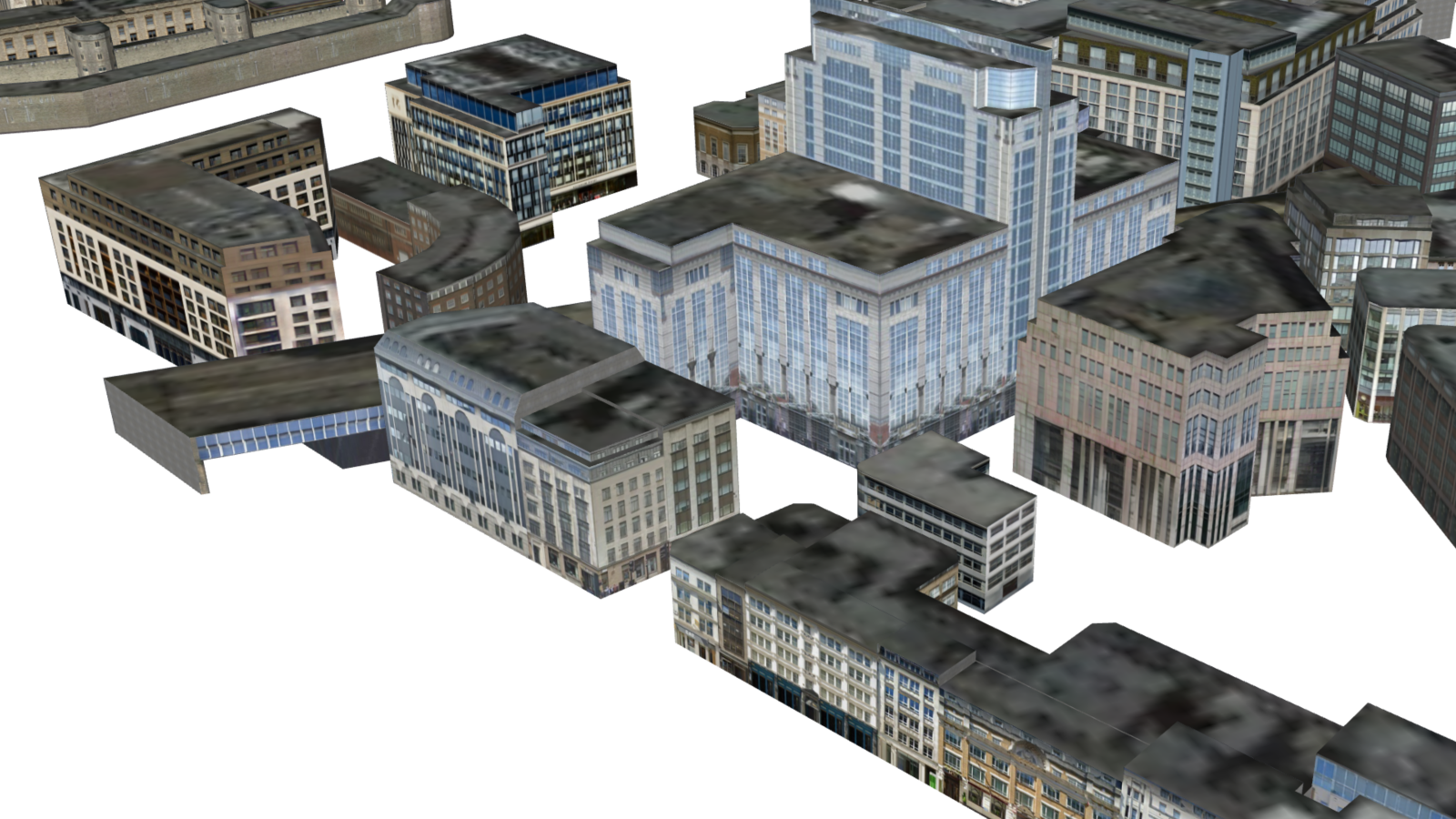}
	}
	\caption{Repaired results of Wellington and London model. (a) Original model of Wellington train station, it has many occlusions on several building fa\c{c}ades. (b) The Wellington train station model after repaired, we de-occluded most occlusions by proposed approach. (c) Original building model which has many missing textures of London. (d) Repaired London model with complete textures synthesized by proposed approach. }
	\label{fig:model_repair}
\end{figure}

\begin{figure}[H]
		\centering
		\includegraphics[width=\linewidth]{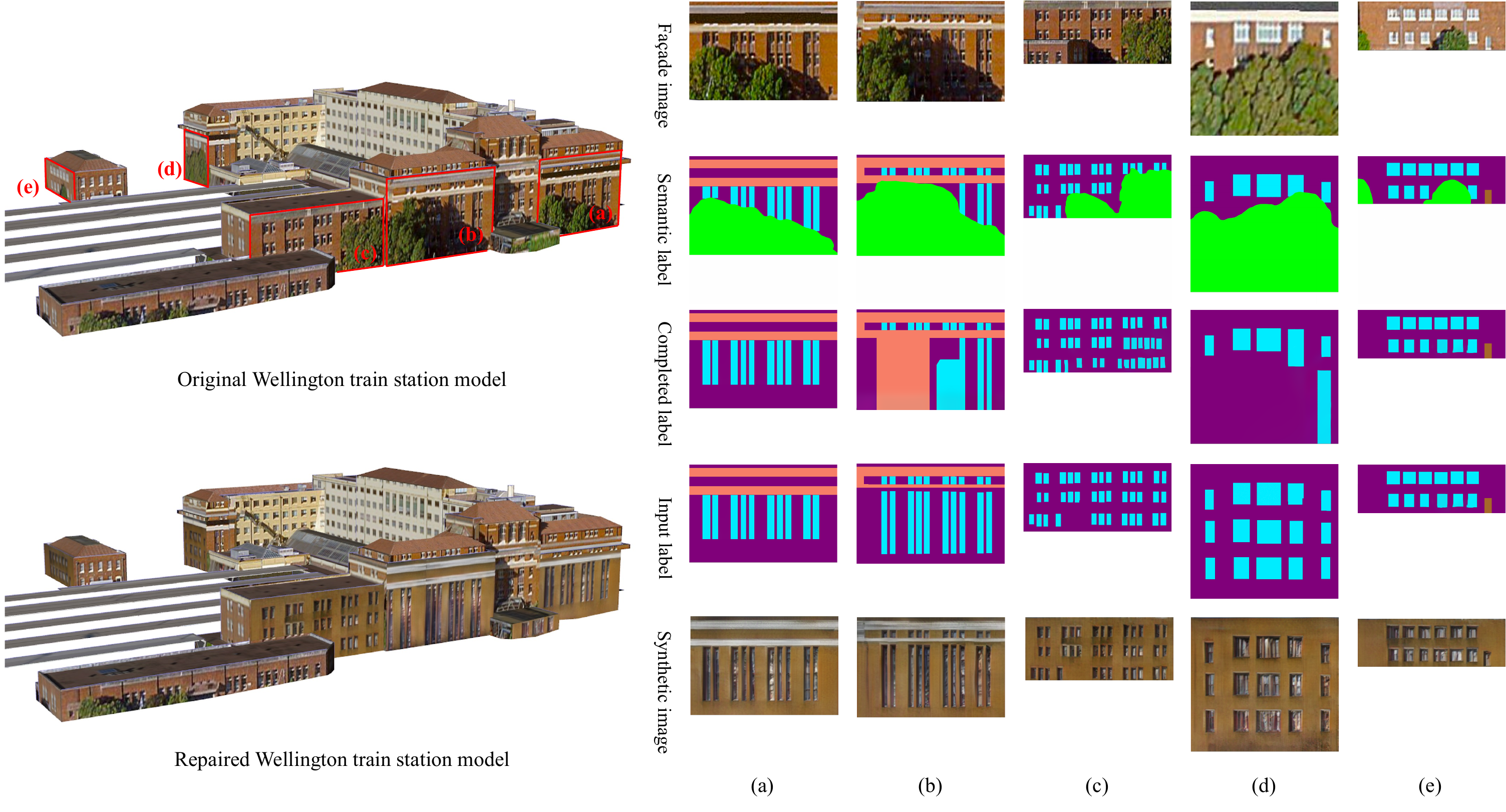}
		\caption{Details of Wellington train station model experiment. All sub-experiments consider (a) as style input. Besides, (a) and (e) use semantics completion results as input labels, others use manual annotated labels as inputs.}
		\label{fig:wellington_repair}
\end{figure}

\begin{figure}[H]
		\centering
		\includegraphics[width=\linewidth]{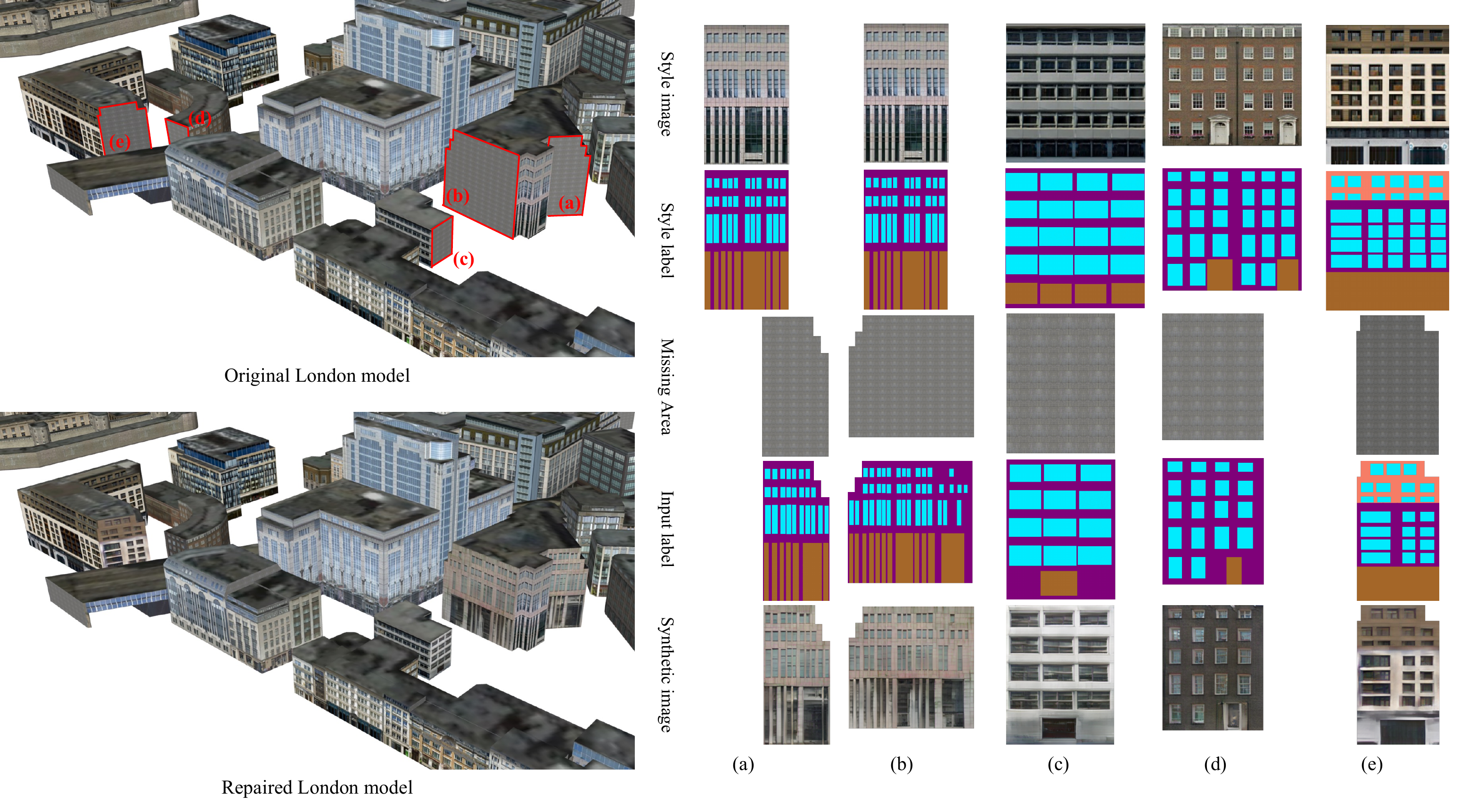}
		\caption{Details of London model experiment. }
		\label{fig:london_repair}
\end{figure}

Figure \ref{fig:model_repair} showcases the building models before and after applying the proposed repair method. The details of the repaired building fa\c{c}ades are presented in Figure \ref{fig:wellington_repair} and \ref{fig:london_repair}. For instance, the Wellington train station building fa\c{c}ade textures have numerous occlusions, resulting in the model having visible defects. To tackle this problem, we first annotate the semantics of the occluded regions and employ the proposed semantics completion method to repair the semantics of those regions. We then manually annotate the semantics of fa\c{c}ades with large occlusions, which can hardly yield satisfactory semantics completion results. Afterwards, we feed the labels recovered manually and automatically and style images with their corresponding semantic labels (\ref{fig:wellington_repair} (a) set in this paper) to the proposed GAN network to synthesize plausible textures. Finally, we map the synthetic images to the respective fa\c{c}ades to obtain the final repaired model.
Moreover, Figure \ref{fig:london_repair} demonstrates the effectiveness of our approach in handling missing textures. We select several buildings in the London dataset with incomplete fa\c{d}ade textures and apply the same process as the Wellington experiment. This experiment also highlights the repair ability of the proposed approach on different architectural styles.

\subsection{Comparison of image translation}
\subsubsection{Qualitative comparison}
\begin{figure}[H]
	\centering
	\subcaptionbox{}[0.16\linewidth]{
		\includegraphics[width=\linewidth]{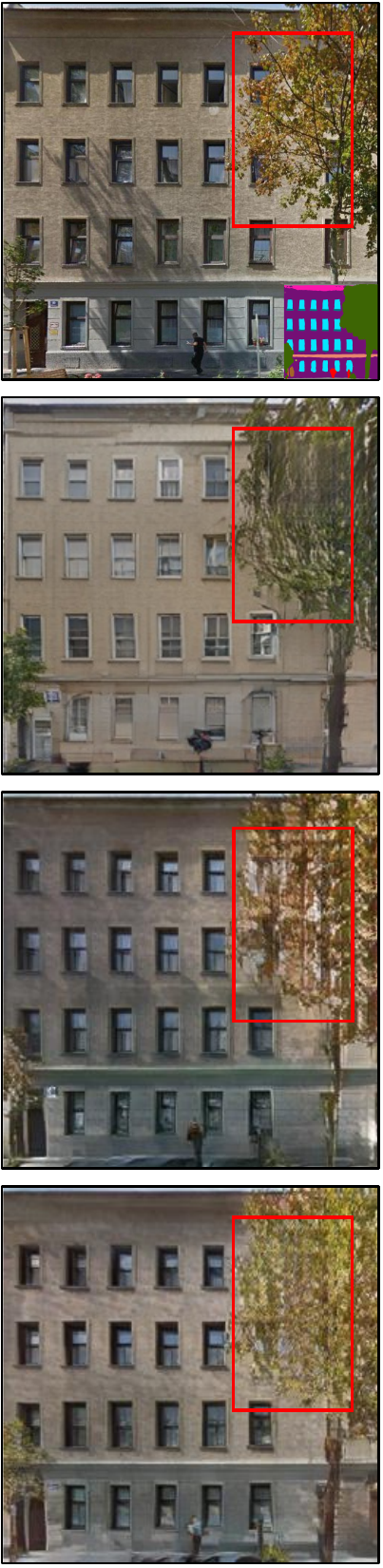}
	}
	\subcaptionbox{}[0.16\linewidth]{
		\includegraphics[width=\linewidth]{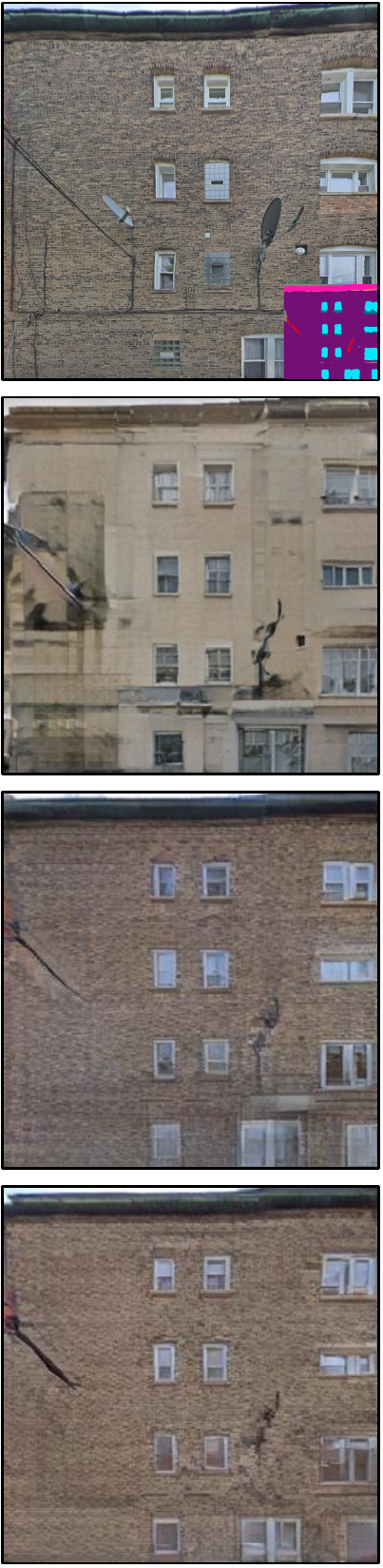}
	}
	\subcaptionbox{}[0.16\linewidth]{
		\includegraphics[width=\linewidth]{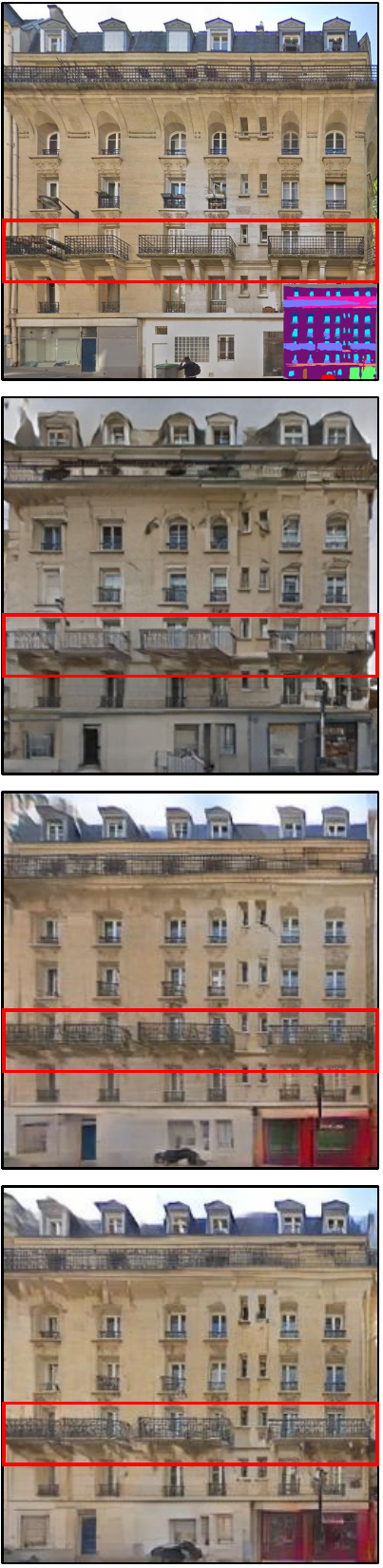}
	}
	\subcaptionbox{}[0.16\linewidth]{
		\includegraphics[width=\linewidth]{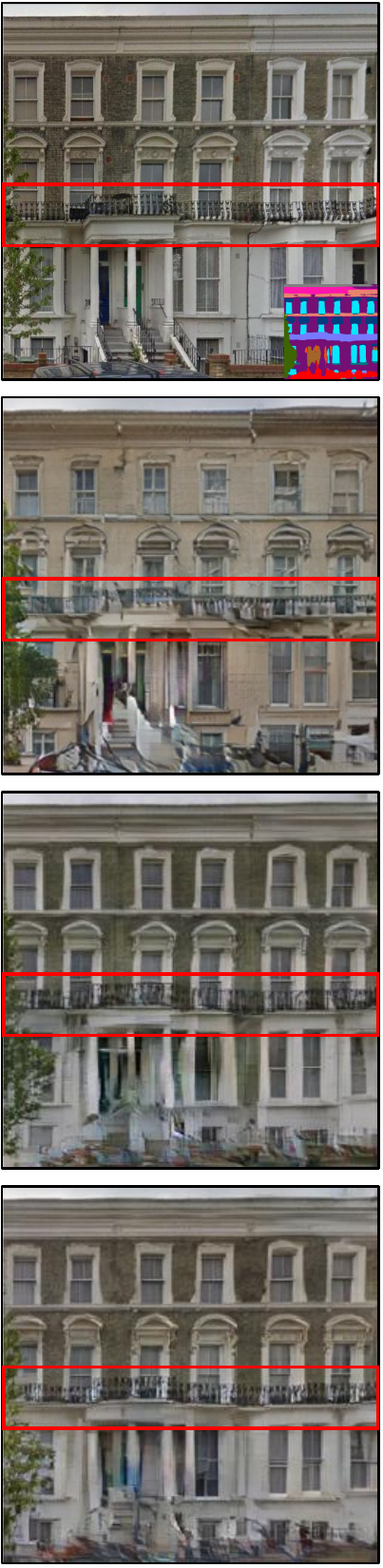}
	}
	\subcaptionbox{}[0.16\linewidth]{
		\includegraphics[width=\linewidth]{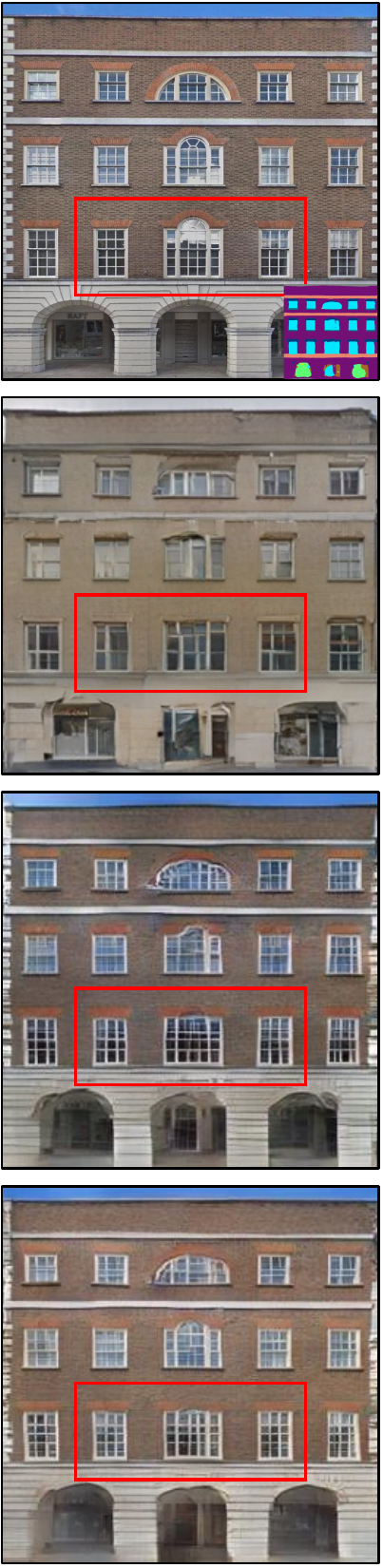}
	}
	\subcaptionbox{}[0.16\linewidth]{
		\includegraphics[width=\linewidth]{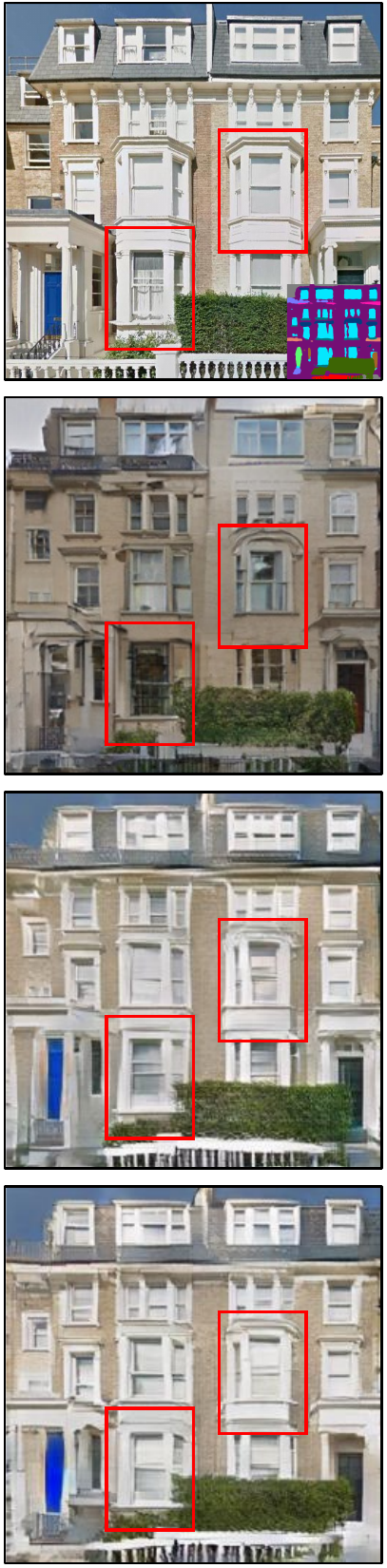}
	}
	
	\subcaptionbox{}[0.16\linewidth]{
		\includegraphics[width=\linewidth]{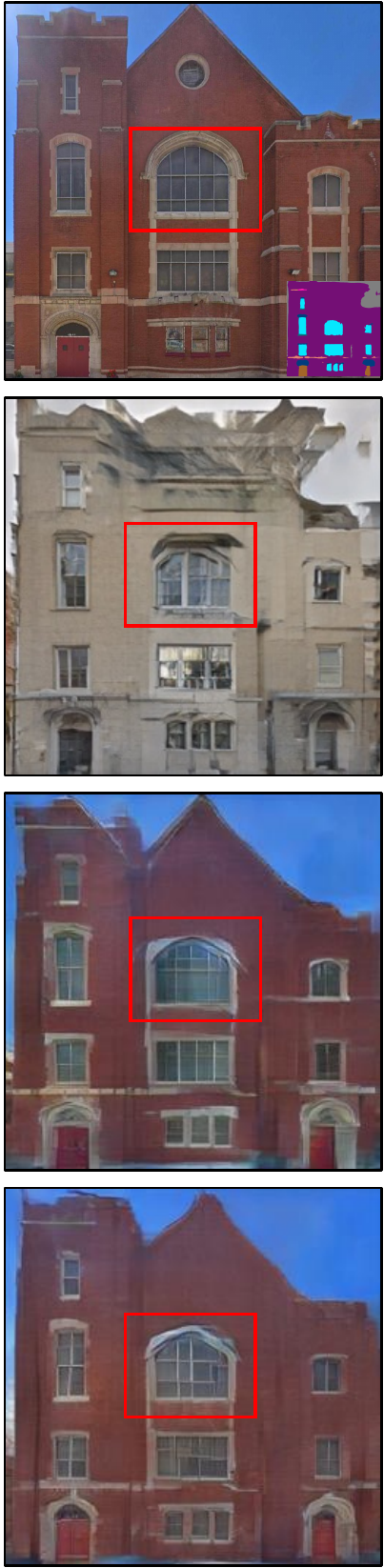}
	}
	\subcaptionbox{}[0.16\linewidth]{
		\includegraphics[width=\linewidth]{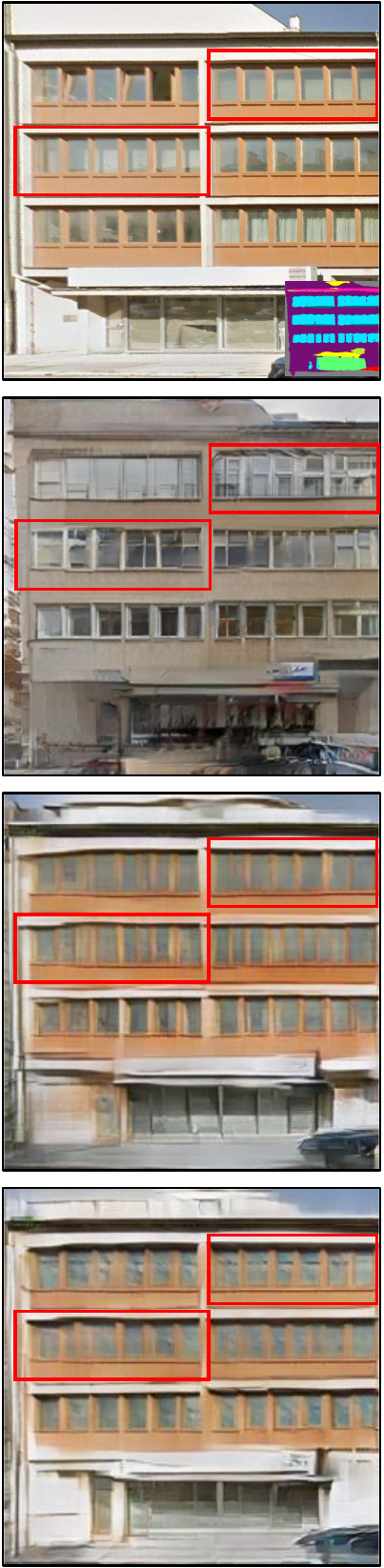}
	}
	\subcaptionbox{}[0.16\linewidth]{
		\includegraphics[width=\linewidth]{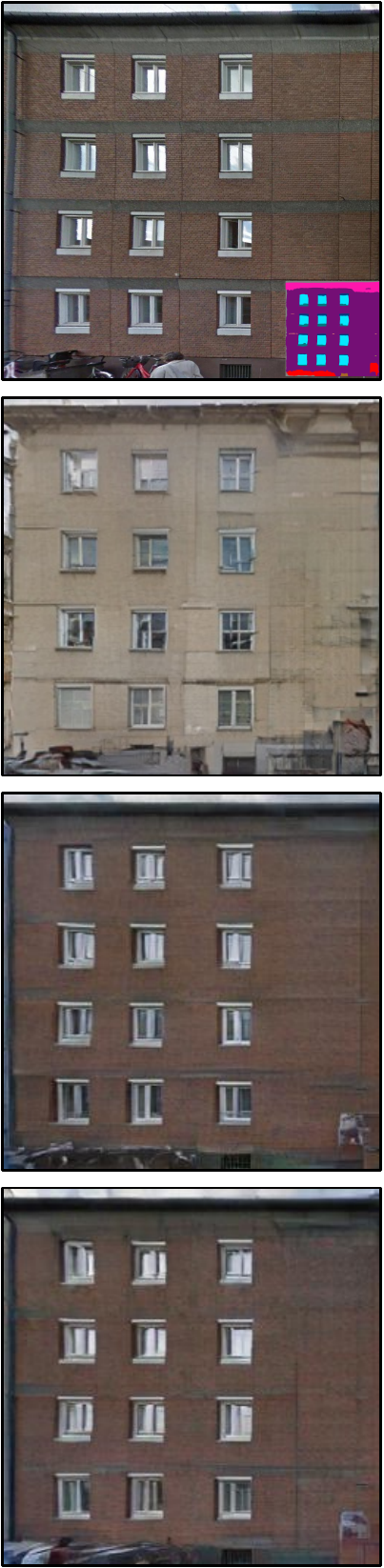}
	}
	\subcaptionbox{}[0.16\linewidth]{
		\includegraphics[width=\linewidth]{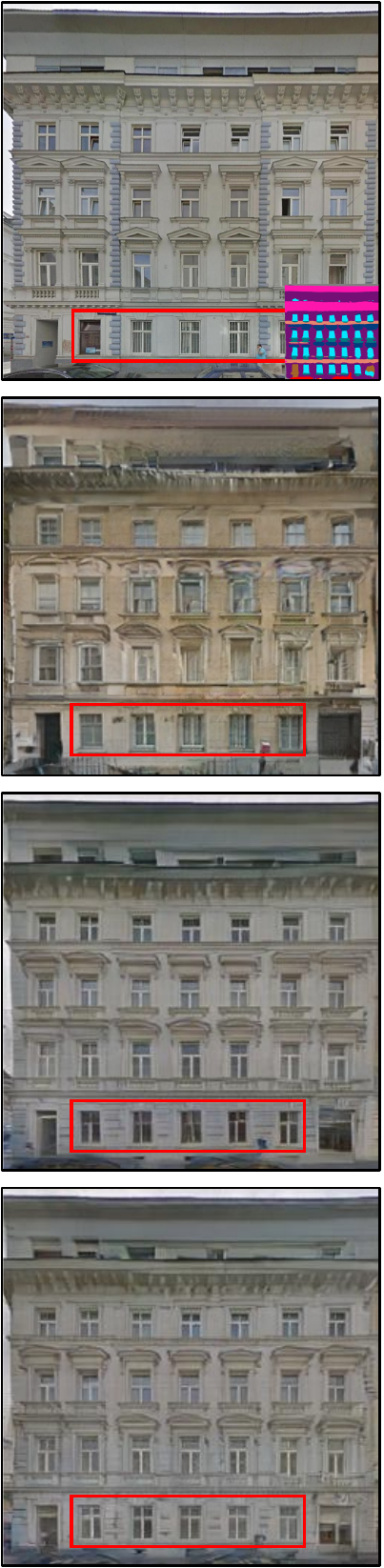}
	}
	\subcaptionbox{}[0.16\linewidth]{
		\includegraphics[width=\linewidth]{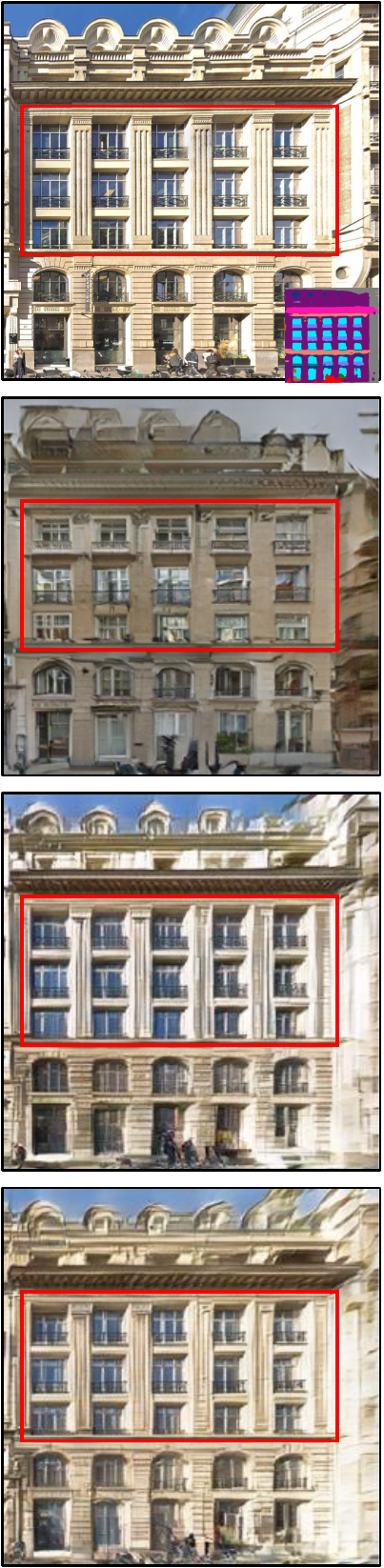}
	}
	\subcaptionbox{}[0.16\linewidth]{
		\includegraphics[width=\linewidth]{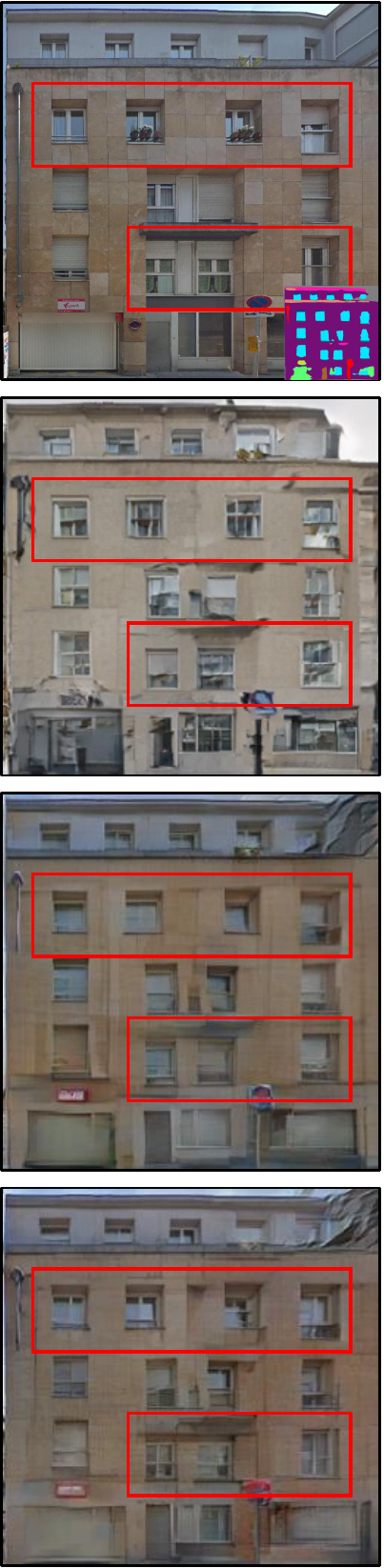}
	}
	\caption{Image translation results of different methods. For each sub-figure, from top to bottom, it is the ground-truth, results of SPADE, results of SEAN and results of proposed method. The red rectangles indicate the areas with differences for synthetic images of different methods.}
	\label{fig:image_translation}
\end{figure}

To evaluate the proposed image translation method, we adopt SPADE, SEAN and our method on LSAA dataset to do a series of experiments. Figure \ref{fig:image_translation} shows some synthesized results of different methods. The first row in every sub-figure denotes the inputs, that specify style and content to synthetic images. The next three rows are the synthesized results of different methods mentioned above, and the red rectangles indicate the obviously differences between different methods. SPADE only considers semantic label map as input and cannot specify the style of result, consequently it is far different from ground-truth. Comparing the results of SEAN and proposed method, it can be seen from Figure \ref{fig:image_translation} (a), (b), (c), (d) that the proposed method has better texture details than SEAN, especially in Figure \ref{fig:image_translation} (c) and (d), the results of proposed method have much more clear and regular details on balustrade areas highlighted by red rectangles. In addition to this, our method has better regular structures generation ability. More specifically, proposed method can synthesizes windows with clear regular structures in Figure \ref{fig:image_translation} (e), (g), (j), (k) and (l), on the contrary, SEAN even cannot generate any structures of windows in the experiments of Figure \ref{fig:image_translation} (j), (k), (l). Besides, it also can be seen from Figure \ref{fig:image_translation} (f), (h) and (i), that the synthesized fa\c{c}ade components of proposed methods are more horizontal and vertical, which is more similar to the ground-truth.

\begin{figure}[H]
	\centering
	\subcaptionbox{Input}[0.24\linewidth]{
		\includegraphics[width=\linewidth]{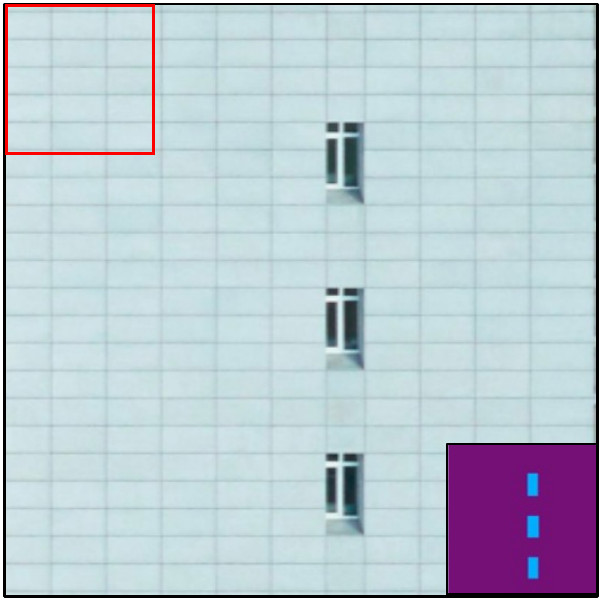}
	}
	\subcaptionbox{Synthesized image}[0.24\linewidth]{
		\includegraphics[width=\linewidth]{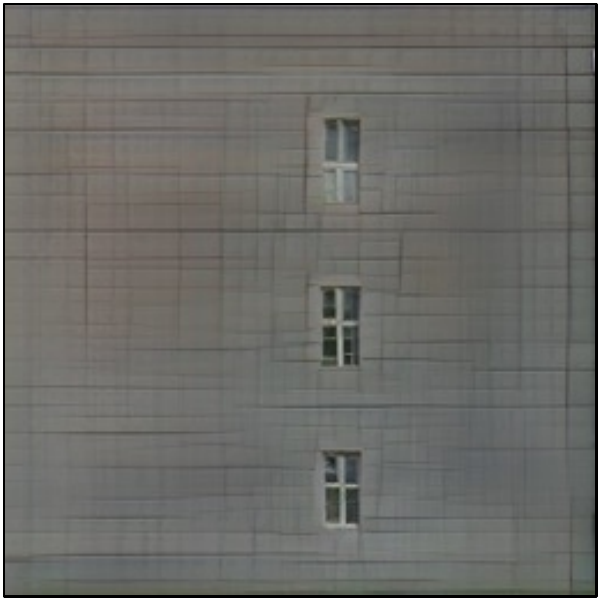}
	}
	\subcaptionbox{Quilted texture}[0.24\linewidth]{
		\includegraphics[width=\linewidth]{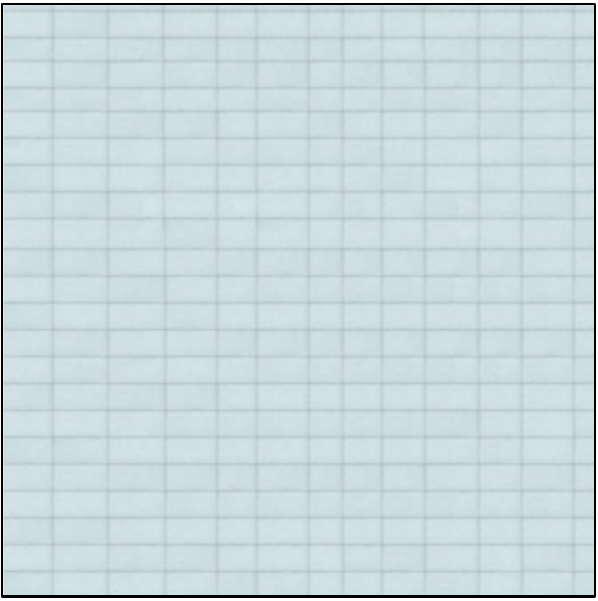}
	}
	\subcaptionbox{Finally result}[0.24\linewidth]{
		\includegraphics[width=\linewidth]{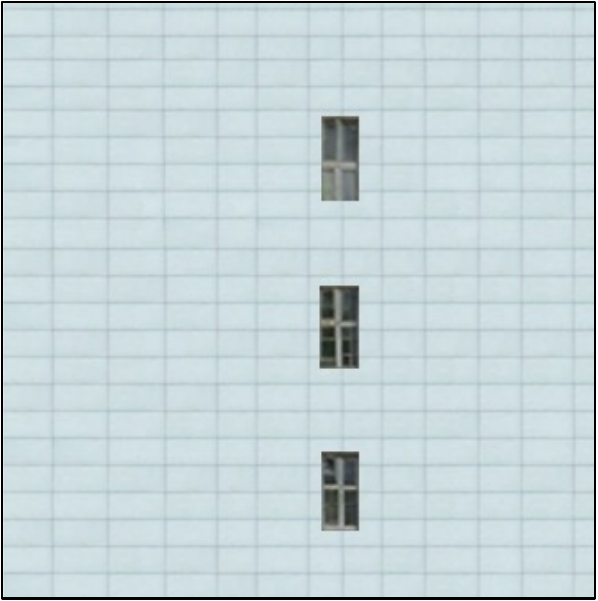}
	}
	\caption{Result after image quilting. (a) Inputs of GAN network, including an ground-truth image with its semantic label map, the red rectangle is the selected input of image quilting algorithm. (b) Synthesized fa\c{c}ade image of proposed network. (c) Synthesized texture of wall region from the pixel patch highlighted in (a). (d) Finally result after replacing wall region with quilted texture. }
	\label{fig:quilting}
\end{figure}

As shown in Figure \ref{fig:quilting}, proposed image translation method still doesn't have enough ability on image synthesis of unseen highly structured styles. Thus, we use image quilting algorithm to eliminate the impact of this deficiency on the results. Figure \ref{fig:quilting} (d) shows the finally result after the replacement of image quilting texture.

\subsubsection{Quantitative comparison}
Peak signal-to-noise ratio (PSNR) and Structural similarity (SSIM)  are wildly used indexes for image quality evaluation. PSNR can be formally expressed as Equation \ref{equ:psnr}.
\begin{equation}
	\label{equ:psnr}
	\left\{\begin{array}{l}
		\text { MSE }= {\lVert \mathbf{R'}-\mathbf{R} \rVert_2} ^{2} / N\\
		\text { PSNR }=10 \log _{10}\left( {{M}^{2} } / {\text { MSE }} \right)
	\end{array}\right.
\end{equation}
where MSE denotes mean square error, $M$ is the range of the data type of $\mathbf{R}$ and $\mathbf{R'}$. SSIM can be calculated by Equation \ref{equ:ssim}.
\begin{equation}
	\label{equ:ssim}
	\text{SSIM}=\frac{1}{N} \sum_{i=1}^{N}\frac{\left(2 \mu_{\mathbf{R'}} \mu_{\mathbf{R}}+C_{1}\right)\left(2 \sigma_{\mathbf{R'} \mathbf{R}}+C_{2}\right)}{\left(\mu_{\mathbf{R'}}^{2}+\mu_{\mathbf{R}}^{2}+C_{1}\right)\left(\sigma_{\mathbf{R'}}^{2}+\sigma_{\mathbf{R}}^{2}+C_{2}\right)}
\end{equation}
where $N$ is the total number of pixels of images $\mathbf{R'}$ and $\mathbf{R}$. $\mu_{\mathbf{R'}}$, $\mu_{\mathbf{R}}$, $\sigma_{\mathbf{R'}}$, $\sigma_{\mathbf{R}}$ and $\sigma_{\mathbf{R'}\mathbf{R}}$ are the local means, standard deviations, and cross-covariance for $\mathbf{R'}$ and $\mathbf{R}$. And in this paper, $C_1 =  (0.01 \times 255) ^ 2$, $C_2 =  (0.03 \times 255) ^ 2$ and $C_3 = (0.03 \times 255) ^ 2 / 2$. SSIM evaluates similarities between two images from luminance,  contrast and structure.

PSNR and SSIM evaluate the differences between synthesized image and ground-truth image from different perspectives. The higher their values are, more similar the synthesized image is to the ground-truth image, which also means that the method is better.

Learned Perceptual Image Patch Similarity (LPIPS), also named perceptual loss, is a metric that measure the differences between two images on feature level by using pre-trained network, e.g. pre-trained AlexNet in this paper. It is more similar to human perception than PSNR and SSIM, and the lower value denotes better synthesis results.

\begin{table}[H]
	\centering
	\caption{Quantitative comparison of 3 different methods on LSAA. The best results are highlighted in bold.}
	\label{tab:quantitative_translation}
	\begin{tabular}{c|ccc}
		\toprule
		\diagbox{Metrics}{Method}  & SPADE & SEAN  & proposed       \\ \midrule
		PSNR $\uparrow$   & 13.77 & 16.76 & \textbf{17.13} \\
		SSIM  $\uparrow$    & 0.291 & 0.489 & \textbf{0.502} \\
		LPIPS  $\downarrow$  & 0.644    & 0.555 & \textbf{0.550} \\ 
		\bottomrule
	\end{tabular}
\end{table}

We trained SPADE, SEAN and proposed method on LSAA training dataset (28494 fa\c{c}ade images with semantic labels) with 4 $\times$ NVIDIA 3090. These trained models are applied to LSAA testing dataset, which contains 1000 fa\c{c}ade images with semantic labels, that never be seen during training. The quantitative results of the comparison are presented in Table \ref{tab:quantitative_translation}, where the proposed method outperformed SPADE and SEAN in all three quality assessment metrics on the LSAA dataset.

\subsection{Analysis of detail and regularity losses}

\begin{figure}[H]
	\centering
	\subcaptionbox{}[0.16\linewidth]{
		\includegraphics[width=\linewidth]{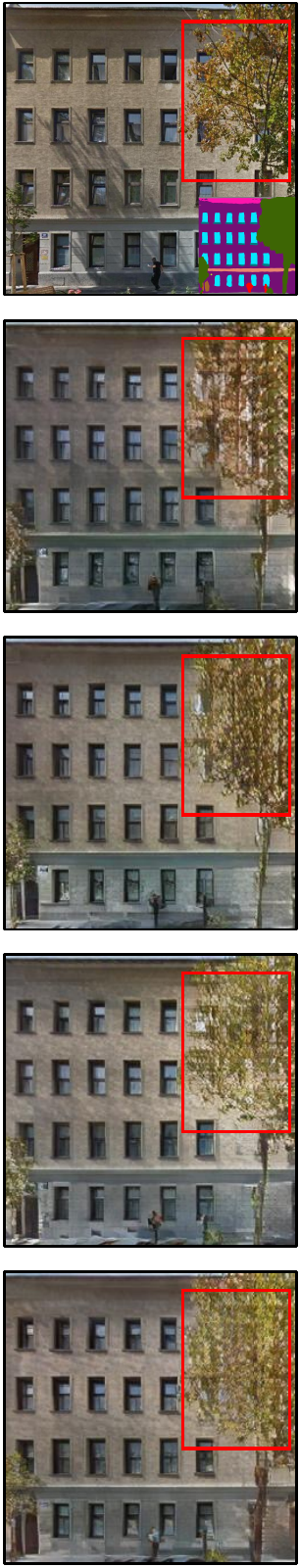}
	}
	\subcaptionbox{}[0.16\linewidth]{
		\includegraphics[width=\linewidth]{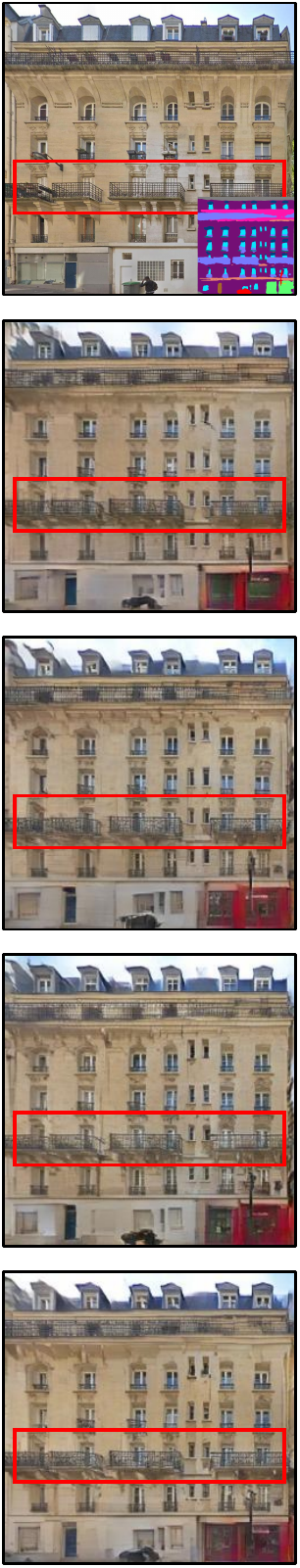}
	}
	\subcaptionbox{}[0.16\linewidth]{
		\includegraphics[width=\linewidth]{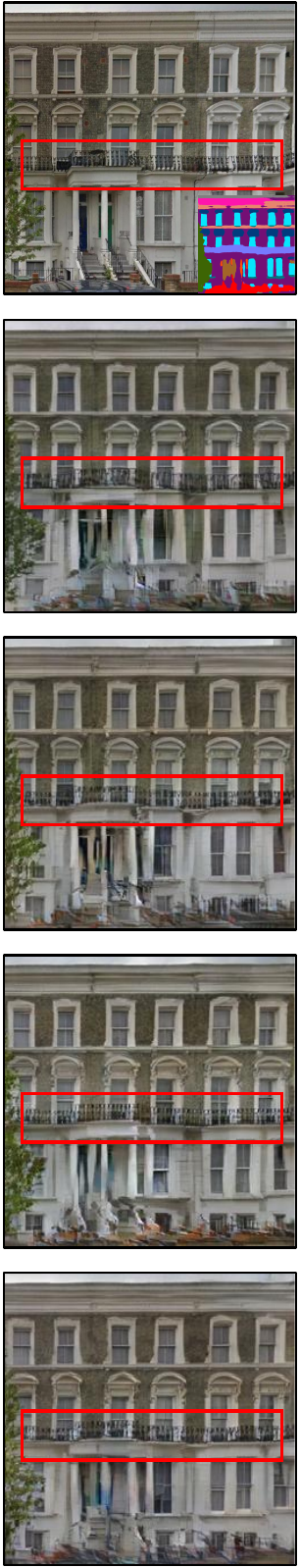}
	}
	\subcaptionbox{}[0.16\linewidth]{
		\includegraphics[width=\linewidth]{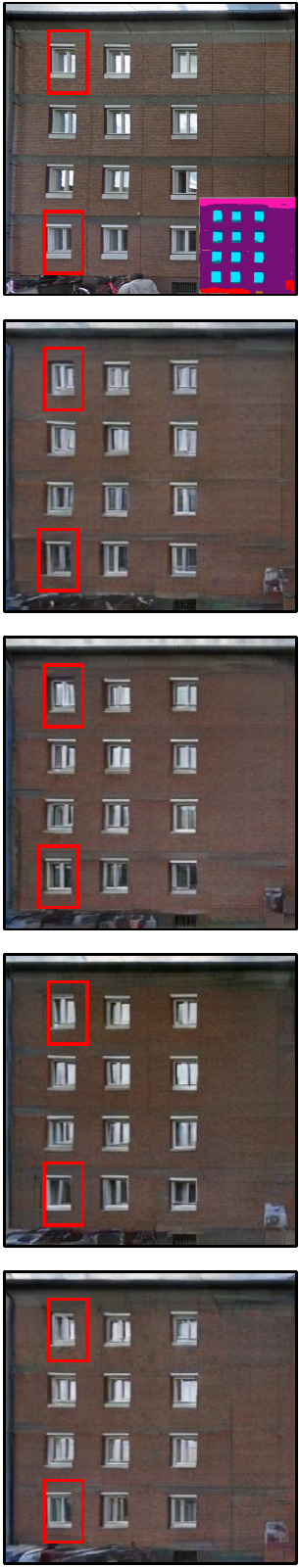}
	}
	\subcaptionbox{}[0.16\linewidth]{
		\includegraphics[width=\linewidth]{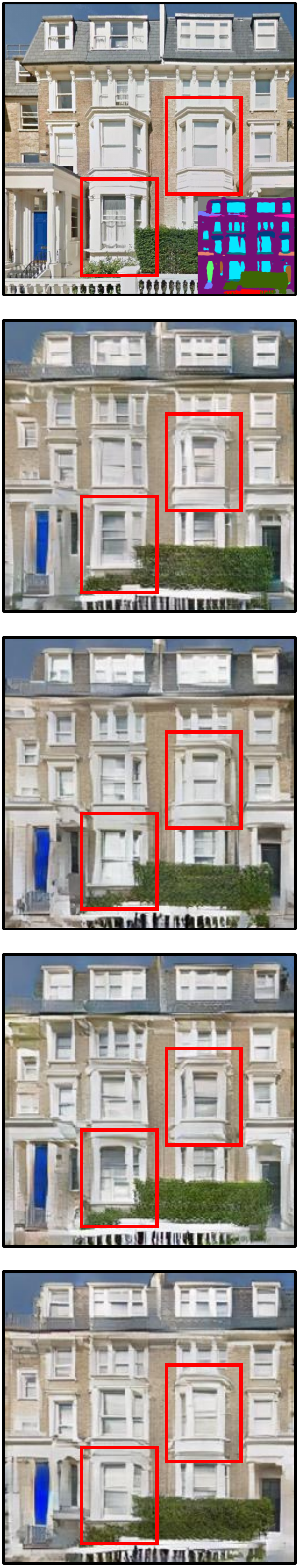}
	}
	\subcaptionbox{}[0.16\linewidth]{
		\includegraphics[width=\linewidth]{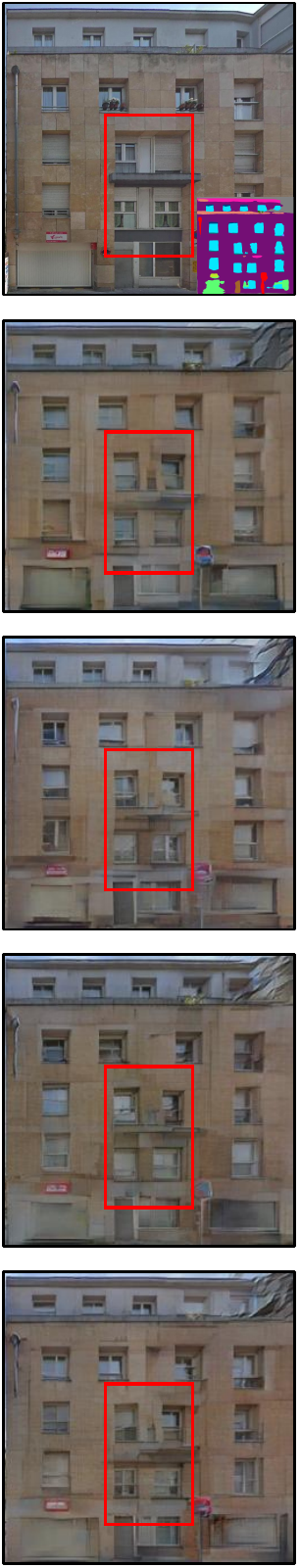}
	}
	\caption{Ablation experiments on LSAA dataset, and the baseline is SEAN. From top to bottom in every sub-figure, it is ground-truth,  w/o. both, w/o. regularity loss, w/o. detail loss, and proposed. Red rectangles indicate regions that have differences with each others.}
	\label{fig:ablation}
\end{figure}

The proposed image translation method enriches details and regularizes the structures of synthesized results by multi-domain losses. In order to prove the availability of our innovation, we conducted ablation experiment on LSAA testing dataset by setting different losses during training. Figure \ref{fig:ablation} shows several typical results synthesized by different settings of proposed method.

Focusing on the red rectangles in Figure \ref{fig:ablation},  it can be concluded that both detail loss and regularity loss proposed by us are effective for the improvement of synthesis results. As we can see the vegetation regions in Figure \ref{fig:ablation} (a) and (e), detail loss can make the synthesized textures be more natural with vivid details. Comparing the balustrade areas in Figure \ref{fig:ablation} (b) and (c), we can find out that the results with detail loss have more clear structures in details. Unfortunately, its regularity maintenance capacity is still not enough for some highly structured fa\c{c}ade components synthesis, and the regularity loss can reinforces it to a certain degree. The window frames in Figure \ref{fig:ablation} (d), (e) and (f) are not plausible enough when synthesized by the model without regularity and detail losses. However, as shown in Figure \ref{fig:ablation} (d), when detail loss is added, the results have distinct difference in window areas. In addition, Figure \ref{fig:ablation} (e), (f) also have significant improvement when combining regularity loss with detail loss.

Table \ref{tab:quantitative_analysis} shows the evaluation of synthesized results under different loss settings by quality assessment indexes. It is demonstrated that detail loss and regularity loss are valid for fa\c{c}ade images synthesis which are always highly structured, and their combination will make the results better and more realistic than their own.

\begin{table}[H]
	\centering
	\caption{Quantitative evaluation of results under different settings on proposed method, baseline is SEAN \citep{zhu2020sean}. The best results are highlighted in bold.}
	\label{tab:quantitative_analysis}
	\begin{tabular}{@{}c|cccc@{}}
		\toprule
		\diagbox{Metrics}{Method}  & w/o. both & w/o. detail loss & w/o. regularity loss & proposed       \\ \midrule
		PSNR  $\uparrow$        & 16.76     & 17.02    & 16.96       & \textbf{17.13} \\
		SSIM $\uparrow$        & 0.489     & 0.500    & 0.496       & \textbf{0.502} \\
		LPIPS $\downarrow$  & 0.555     & 0.551    & 0.553       & \textbf{0.550} \\ \bottomrule
	\end{tabular}
\end{table}

\subsection{Discussion and limitations}

The proposed approach for repairing defective building fa\c{c}ade textures yields plausible results, addressing the problem of texture occlusion or missing in 3D realistic building models. The repaired models exhibit reasonable textures and can be used for model exhibition. The proposed semantics completion method also produces desirable results for occlusions, improving the automation of the texture repair approach. Moreover, the regularity-aware multi-domain universal image translation method has demonstrated the ability to synthesize more detailed and structured building fa\c{c}ade textures through qualitative experiments. Quantitative comparisons also show the superiority of the proposed image translation method compared to others.

Despite the better results of fa\c{c}ade texture repair and image translation, there are still limitations. The resolution and definition of style images affect the final synthesis results. For higher resolution, we can resolve this problem by simple down-sampling. However, for lower resolution, it is hard to realize a plausible result by up-sampling. In addition to this, GAN based methods are hardly to train and too computation-consuming to synthesis higher resolution images. Future research will explore the combination of image translation and super-resolution techniques.

\section{CONCLUSION}
\label{s:conclusion}

3D building model is the fundamental of digital city, live navigation and smart driving, its realism usually comes from photogrammetric textures. However, there are inevitable occlusions in built-up areas, the vegetation near buildings also makes it difficult to acquire un-occluded fa\c{c}ade textures by handheld camera. The above problems eventually lead to the inaccessible areas of photogrammetry, which is cannot deal with in traditional texture mapping pipeline. To solve this problem, this paper proposed a deep learning based approach to repair fa\c{c}ade defect textures from easily accessible semantic label map. Specifically, we proposed a semantics recovery method by using image completion algorithm to improve automation of de-occlusion. A regularity-aware multi-domain universal image translation method is used to synthesize building fa\c{c}ade textures of arbitrary styles. This method achieves better results by enriching the details and improving the regularity of synthesized images. Overall, the proposed texture repair approach not only can de-occlusion, but also can generate realistic fa\c{c}ade textures which have actual architectural style from nothing. Future directions on the building models processing may include: (1) de-occlusion of other regions by proposed approach; (2) high-resolution texture synthesis under limited computation resources; (3) deep learning based geometry generation of 3D model.

\section*{Acknowledgments}
This work was supported in part by the National Natural Science Foundation of China (Project No. 42230102, 42071355, 41871291) and the National Key Research and Development Program of China (Project No. 2022YFF0904400).

\bibliographystyle{model2-names}
\bibliography{facadetexture}

\end{document}